\RequirePackage{fix-cm}
\documentclass[twocolumn, natbib]{svjour3}          
\smartqed  
\usepackage{graphicx}
\usepackage{amsmath}
\usepackage{amssymb}
\usepackage{bbm}
\usepackage{multirow}
\usepackage{booktabs}
\usepackage{url}
\usepackage[draft,colorlinks=true,pagebackref=true,breaklinks=true,bookmarks=false]{hyperref} 

\usepackage{paralist}
\usepackage[normalem]{ulem}

\newcommand{\abc}[1]{\textcolor{black}{#1}}

\newcommand{\zq}[1]{\textcolor{black}{#1}} 

\newcommand{\add}[1]{\textcolor{black}{#1}} 

\newcommand{\zqq}[1]{\textcolor{black}{#1}} 
\newcommand{\zqqblue}[1]{\textcolor{black}{#1}} 

\newcommand{\abcn}[1]{\textcolor{black}{#1}}
\newcommand{\nabc}[1]{\textcolor{black}{#1}} 
\newcommand{\nnabc}[1]{\textcolor{black}{#1}} 

\newcommand{\sssection}[1]{{\em{#1}}}

\graphicspath{{figures/}}

\journalname{xxx}

\begin{document}

\title{Wide-Area Crowd Counting: Multi-View Fusion Networks for Counting in Large Scenes
}


\author{Qi Zhang         \and
        Antoni B. Chan 
}


\institute{Qi Zhang \at
              City University of Hong Kong \\
              \email{qzhang364-c@my.cityu.edu.hk}           
           \and
           Antoni B. Chan \at
              City University of Hong Kong \\
              \email{abchan@cityu.edu.hk}           
}

\date{Received: date / Accepted: date}

\maketitle

\begin{abstract}
  Crowd counting in single-view images has achieved outstanding performance on existing counting datasets. However, single-view counting is not applicable to large and wide scenes (\emph{e.g.}, public parks, long subway platforms, or event spaces) because a single camera cannot capture the whole scene in adequate detail for counting, \emph{e.g.}, when the scene is too large to fit into the field-of-view of the camera, too long so that the resolution is too low on faraway crowds, or when there are too many large objects that occlude large portions of the crowd. Therefore, to solve the wide-area counting task requires multiple cameras with overlapping fields-of-view. In this paper, we propose a deep neural network framework for multi-view crowd counting, which fuses information from multiple camera views to predict a scene-level density map on the ground-plane of the 3D world. We consider three versions of the fusion framework: the late fusion model fuses camera-view density map; the na\"ive early fusion model fuses camera-view feature maps; and the multi-view multi-scale early fusion model ensures that features aligned to the same ground-plane point have consistent scales. A rotation selection module further ensures consistent rotation alignment of the features. We test our 3 fusion models on 3 multi-view counting datasets, PETS2009, DukeMTMC, and a newly collected multi-view counting dataset containing a crowded street intersection. Our methods achieve state-of-the-art results compared to other multi-view counting baselines.

\end{abstract}

\section{Introduction}

\begin{figure}
   \includegraphics[width=\linewidth]{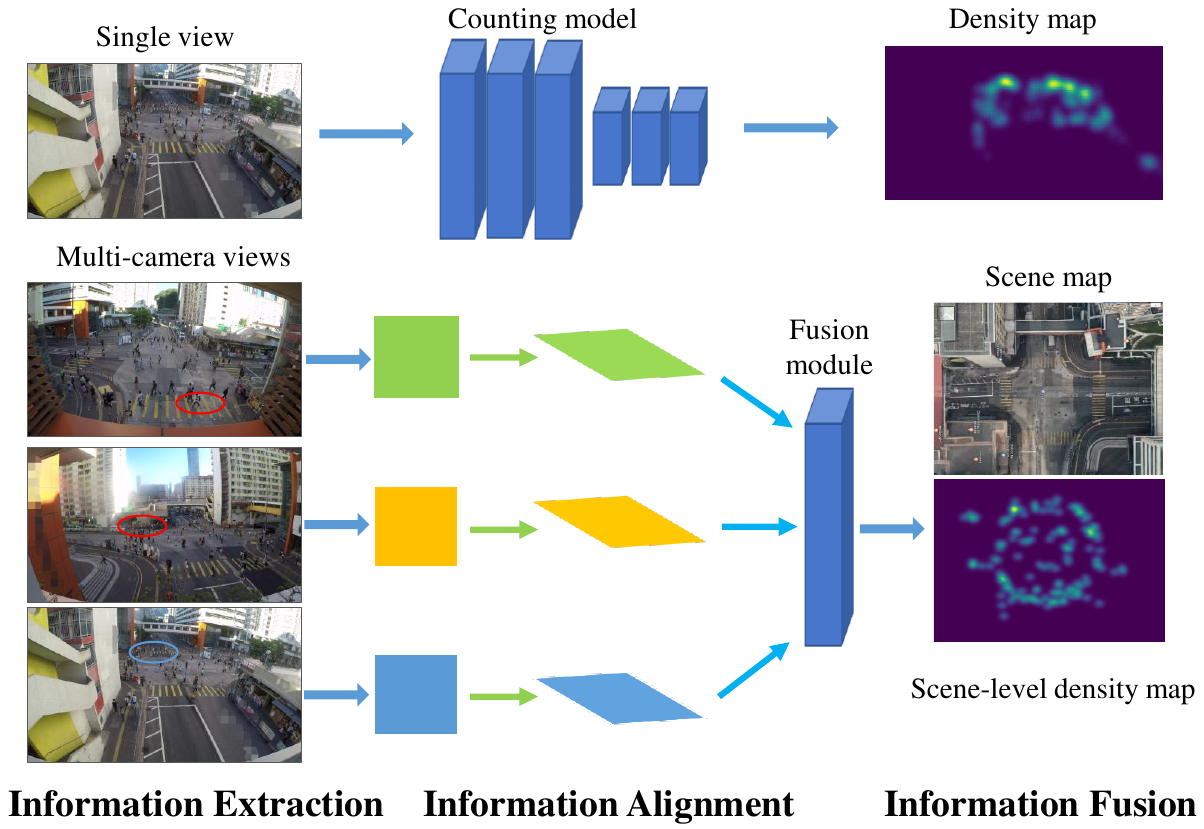}
   \caption{The pipeline of the proposed multi-view fusion framework comparing with the single image counting framework. In the multi-view fusion model, feature maps are extracted from multiple camera views, aligned on the ground-plane, and fused to obtain the scene-level ground-plane density map.
   \abc{The scene map is shown for reference.}
   \zqq{For single image counting, many people are occluded (in red circles) and in low resolution (in blue circles), which decreases the counting performance.}
   }
\label{fig:abstract_pipepline}
\end{figure}

\begin{figure*}
\begin{center}
   \includegraphics[width=0.9\linewidth]{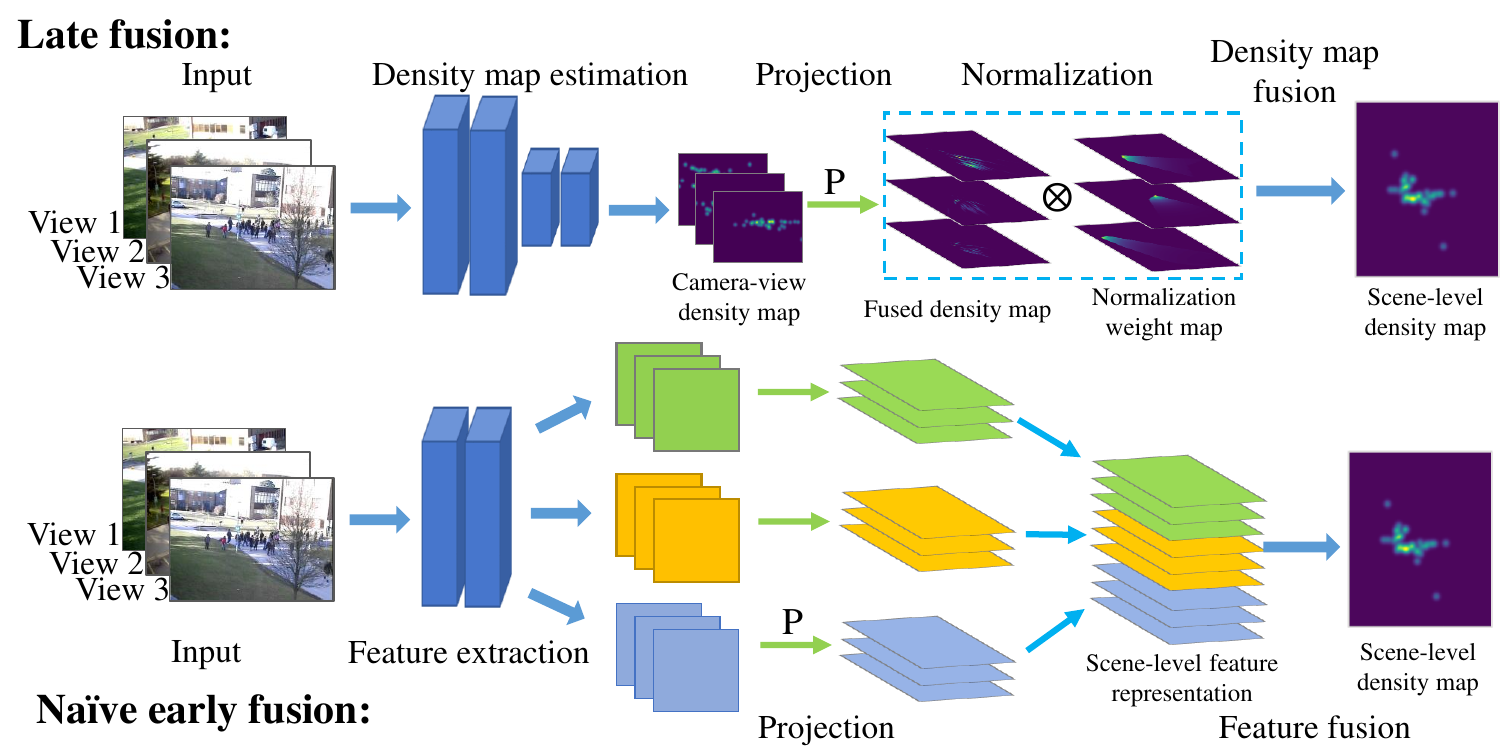}
\end{center}
   \caption{The pipeline of our late fusion model and na\"ive early fusion model for multi-view counting. In the late fusion model, single-view density maps are fused. In the na\"ive early fusion model, single-view feature maps are fused.
   }
\label{fig:two_models}
\end{figure*}

\par Crowd counting aims to estimate the number of the people in images or videos. It has a wide range of real-world applications, such as crowd management, public safety, traffic monitoring or urban planning \citep{sindagi2018survey}. For example, crowd counting can detect overcrowding on the railway platform and help with the train schedule planning. Furthermore, the estimated crowd density map provides spatial information of the crowd, which can benefit other tasks, such as human detection \citep{eiselein2013enhancing, kang2018beyond, ma2015small} and tracking \citep{kang2018beyond, ren2018fusing, rodriguez2011density}.

\par Recently, with the strong learning ability of deep neural networks (DNNs), density map based crowd counting methods have achieved outstanding performance on the existing counting datasets \citep{cao2018scale, idrees2018composition, sindagi2017generating}, where the goal is to count the crowd in a single image. However, a single image view is not adequate to cover a \emph{large} and \emph{wide} scene, such as a large park or a long train platform. For these wide-area scenes, a single camera view cannot capture the whole scene in adequate detail for counting, either because the scene is too large (wide) to fit within the field-of-view of the camera, or the scene is too long so that the resolution is too low in faraway regions. Furthermore, a single view cannot count regions that are still within the scene, but are totally occluded by large objects (\emph{e.g.}, trees, large vehicles, building structures). Therefore, to solve the wide-area counting task requires multiple camera views with overlapping field-of-views, which combined can cover the whole scene and can see around occlusions. The goal of wide-area counting is then to use multiple camera views to estimate the crowd count of the whole scene.

\par Existing multi-view counting methods rely on foreground extraction techniques and hand-crafted features. Their crowd counting performance is limited by the effectiveness of the foreground extraction, as well as the representation ability of hand-crafted features. Considering the strong learning power of DNNs as well as the performance progress of single view counting methods using density maps, the feasibility of end-to-end DNNs-based multi-view counting methods should be explored.

\par In this paper, we propose a DNNs-based multi-view counting method that extracts information from each camera view and then fuses them together to estimate a scene-level ground-plane density map (see Fig.~\ref{fig:abstract_pipepline}). The method consists of 3 stages: 1) \emph{Information extraction} -- single view feature maps are extracted from each camera image with DNNs; 2) \emph{Information alignment} -- using the camera geometry, the feature maps from all cameras are projected onto the ground-plane in the 3D world so that the same person's features are approximately aligned across multiple views, and properly normalized to remove projection effects; 3) \emph{Information fusion} -- the aligned single-view projected feature maps are fused together and used to predict the scene-level ground-plane density map.

\par
\zqqblue{
As single-view crowd counting is relatively mature, it has well-studied feature extractor and decoder architectures for predicting camera-level density maps.
Building from this, multi-view crowd counting can leverage single-view feature extractors and predicted density maps for each camera-view.
The key issue then is \emph{what} and \emph{how} to fuse the information from the various cameras into a ground-plane representation for decoding into a ground-plane density map.
We consider three variants of fusion to provide a thorough study on the fusion architecture for multi-view counting DNNs models. These three variants differ in \emph{what information is fused} (i.e., single-view density maps or feature maps) and \emph{how fusion occurs} (i.e., simple concatenation or scale/rotation-aware concatenation).  Specifically, the three variants are:
1) concatenation of single-view density maps (denoted as late fusion);
2) concatenation of single-view feature maps (na\"ive early fusion);
3) scale-aware and rotation-aware concatenation of feature maps (multi-view multi-scale, MVMS/MVMSR).
}

Specifically,
first, in our late-fusion model (see Fig.~\ref{fig:two_models} top), view-level density maps are predicted for each camera view, projected to the ground-plane, and fused for estimating the scene-level density map.
\zqqblue{This model fuses count-level information, similar to traditional count-based methods \citep{dittrich2017people, li2012people, ma2012reliable, Maddalena2014people}}.
We also propose a post-projection normalization method that removes the projection effect that distorts the sum of the density maps (and thus the count). Second, in our na\"ive early fusion model (see Fig.~\ref{fig:two_models} bottom), convolutional feature maps are extracted from each camera view, projected to the ground-plane and fused to predict the scene-level density map. Third, to handle the scale variations of the same person across camera views, our multi-view multi-scale (MVMS) early fusion model (see Fig.~\ref{fig:MVMS_model}) extracts features with consistent scale across corresponding locations in the camera views before applying projection and fusion. We consider 2 approaches for selecting the suitable scales, based on distances computed from the camera geometry. To further improve the multi-view fusion performance, a rotation selection module is added in the multi-view fusion step (denoted as MVMSR).

\par The existing multi-view datasets that can be used for multi-view counting are PETS2009 \citep{ferryman2009pets2009} and DukeMTMC \citep{ristani2016MTMC}. However, PETS2009 is not a wide-area scene as it focuses on one walkway, while DukeMTMC is a wide-area scene but does not contain large crowds. To address these shortcomings, we collect a new wide-area dataset from a busy street intersection, which contains large crowds, more occlusion patterns (\emph{e.g.}, buses and cars), and large scale variations. This new dataset more effectively tests multi-view crowd counting in a real-world scene.

\par In summary, our main contributions are:
\begin{enumerate} 
  \item We propose an end-to-end trainable DNNs-based multi-view crowd counting framework, which fuses information from multiple camera views to obtain a scene-level density map.
  \item We propose 3 fusion models based on our multi-view framework (late fusion, na\"ive early fusion, and multi-view multi-scale early fusion), which achieve better counting accuracy compared to baselines.
  \item \add{We propose a rotation selection module based on rotation equivariant networks to further improve the multi-view fusion by considering the geometric properties of the average-height projection.}
  \item We collect a real-world wide-area counting dataset consisting of multiple camera views, which will advance research on multi-view wide-area counting.
\end{enumerate} 

\par The remainder of this paper is organized as follows. In Section 2, existing single-view and multi-view counting methods are reviewed, and the rotation neural networks are introduced. In Section 3, the proposed two DNNs-based multi-view counting models (both late fusion and na\"ive early fusion model) are presented.  In Section 4, the multi-view multi-scale early fusion model with scale selection and rotation selection module is presented. In Section 5, we conduct experiments on multi-view counting datasets.

\section{Related Work}
\par In this section, we review methods for crowd counting from single-view and multi-view cameras, as well as rotation equivariant/invariant networks. 

\subsection{Single-view counting}

\sssection{Traditional methods.}~
Traditional single-view counting methods can be divided into 3 categories \citep{Chen2013Crowd, sindagi2018survey}: detection, regression, and density map methods. Detection methods try to detect each person in the images by extracting hand-crafted features \citep{viola2004robust, sabzmeydani2007detecting, wu2007detection} and then training a classifier \citep{joachims1998text, viola2005detecting, gall2011hough} using the extracted features. However, the detection methods do not perform well when the people are heavily occluded, which limits their application scenarios. Regression methods extract image features \citep{chan2008privacy, cheng2014recognizing, junior2010crowd, krizhevsky2012imagenet} and learn a mapping directly to the crowd count \citep{chan2012counting,chen2012feature, paragios2001mrf, marana1998efficacy}. However, their performance is limited by the weak representation power of the hand-crafted low-level features. Instead of directly obtaining the counting number, \cite{lempitsky2010learning} proposed to estimate density maps, where each pixel in the image contains the local crowd density, and the count is obtained by summing over the density map. Traditional density map methods learn the mapping between the hand-crafted local features and the density maps \citep{lempitsky2010learning, pham2015count, wang2016fast, xu2016crowd}.

\sssection{DNNs-based methods.} DNNs-based crowd counting has mainly focused on density map estimation. The first networks used a standard CNN \citep{zhang2015cross} to directly estimate the density map from an image.
Scale variation is a critical issue in crowd counting, due to perspective effects in the image \citep{Yan_2019_ICCV, Liu_2019_ICCV, Xu_2019_ICCV}. \cite{zhang2016single} proposed the multi-column CNN (MCNN) consisting of 3 columns of different receptive field sizes, which can model people of different scales. \cite{sam2017switching} added a switching module in the MCNN structure to choose the optimal column to match the scale of each patch.
\cite{onoro2016towards} proposed to use the patch pyramid as input to extract multi-scale features. Similarly, \cite{Kang2018Crowd} used an image pyramid with a scale-selecting attention block to adaptively fuse predictions on different scales.

\par Recently, more sophisticated network structures have been proposed and extra information is explored to advance the counting performance
\citep{shi2018crowd, idrees2018composition, Wang2019Learning, ranjan2018iterative, cao2018scale, li2018csrnet,
liu2018decidenet, shen2018crowd, Jiang2019Crowd, Liu2019Context}.
\cite{sindagi2017generating} incorporated global and local context information in the crowd counting framework, and proposed the contextual pyramid CNN (CP-CNN).
\cite{idrees2018composition} proposed composition loss, implemented through multiple dense blocks after branching off the base networks.
\cite{li2018csrnet} replaced pooling operations in the CNN layers with dilated kernels to deliver larger reception fields and achieved better counting performance.
\cite{kang2017incorporating} proposed an adaptive convolution neural network (ACNN) that uses side information (camera angle and height) to include context into the counting framework.
\zqq{Many methods have focused on the perspective change issue in the counting task.
\cite{cao2018scale} extracted multi-scale features with a scale aggregation module and generated high-resolution density maps by using a set of transposed convolutions.
\citet{shi2019revisiting} proposed to estimate the perspective map and use it to adaptively fuse the multi-scale output density maps.
\citet{yan2019perspective} proposed perspective-guided convolution (PGC) to utilize perspective information instead of multi-scale or multi-column architectures.
\citet{yang2020reverse} proposed to estimate a perspective factor to warp the input images to correct the perspective distortions.
}
\cite{Lian2019Density} proposed a regression guided detection network (RDNet) for RGB-D crowd counting.
\cite{Liu2019Recurrent} proposed Recurrent Attentive Zooming Network to zoom high density regions for higher-precision counting and localization.

\par All these methods are using DNNs to estimate a density map \abc{on the image plane} of a single camera-view, with different architectures improving the performance across scenes and views. In contrast, in this paper, we focus on fusing multiple camera views of the same scene to obtain a ground-plane density map in the 3D world. \nabc{These single-view methods serve as the backbone single-view feature extractors for our multi-view fusion networks.}

\subsection{Multi-view counting}
\par Existing multi-view counting methods can be divided into 3 categories: detection/tracking, regression, and 3D cylinder methods. The detection/tracking methods first perform detection or tracking on each scene and obtain single-view detection results. Then, the detection results from each view are integrated by projecting the single-view results to a common \abc{coordinate system}, \emph{e.g.}, the ground plane or a reference view. The count of the scene is obtained by solving a correspondence problem \citep{dittrich2017people, li2012people, ma2012reliable, Maddalena2014people}.
Regression based methods first extract foreground segments \abc{from each view}, then build the mapping relationship of the segments and the count number with a regression model \citep{Ryan2014Scene, Tang2014Cross}. 3D cylinder-based methods try to find the people's locations in the 3D scene by minimizing the gap between the people's 3D positions projected into the camera view and the single view detection \citep{Ge2010Crowd}.

\par These multi-view counting methods are mainly based on hand-crafted low-level features and regression or detection/tracking frameworks. Regression-based methods only give the global count, while detection/tracking methods cannot cope well with occlusions when the scene is very crowded. In contrast to these works, our approach is based on predicting the ground-plane density map in the 3D world by fusing the information across camera views using DNNs. Two advantages of our approach are the abilities to learn the feature extractors and fusion stage in end-to-end training, and to estimate the spatial arrangement of the crowd on the ground plane. While the previous methods are mainly tested on PETS2009, which only contains low/moderate crowd numbers on a walkway, here we test on a newly collected dataset comprising a real-world scene of a street intersection with large crowd numbers, vehicles, and occlusions.

\par A preliminary conference version of this work appears in \cite{zhang2019wide}.
This journal version contains the following extensions: 1)
more details about the scale selection module are added, specifically the rationale of the multi-view scale selection guided by the distance map;
2) a new rotation selection module is proposed to consider the stretching effect of the fixed average-height projection, which further boosts the performance compared to the model in \cite{zhang2019wide};
3) 
\zqq{more experiments and ablation studies, including more evaluation metrics (MAE, MSE, NAE and GAME), experiments of new and updated comparison methods (`feature concatenation', `stitching', and updated 'Detection+ReID' method),
experiments showing how multi-cameras improve single-view counting performance for each multi-camera counting methods,
more ablation studies on the rotation modules (filter number, layer number and quantization angle),
experiments with more recent backbone networks,
comparison results with different module settings and methods and on another test set in DukeMTMC,
and the running speed comparison of different methods.}

\zqq{Finally, following our conference paper (Zhang and Chan 2019), a subsequent work \citep{zheng2021learning} enhances the late fusion model's performance by modeling the correlation between each pair of views for cross view fusion.}


\subsection{Rotation equivariant/invariant networks}
Rotation equivariance or invariance relates to the DNNs' robustness to rotation changes \nabc{of the input image}. Rotation equivariance means the output is accordingly rotated if the input is rotated, which is useful for the dense prediction tasks, like semantic segmentation or density map estimations. Rotation invariance means the output is invariant no matter how the input is rotated, which is useful for classification tasks.

\par
To enhance the networks' robustness to rotations, the easiest method is to use the data augmentation, namely rotating the original examples multiple times and training the network on the rotated versions.
\cite{Jaderberg2015Spatial} introduced the Spatial Transformer which can spatially manipulate data within the network, giving neural networks the ability to actively spatially transform feature maps.
Rotation equivariant and invariant networks have also been proposed to improve the rotation robustness.
\cite{laptev2016ti} uses multiple rotated examples as inputs into a shared network to extract features at multiple rotations, and then uses a max-pooling layer among these features to obtain rotation-invariant features.
In addition to image rotating, feature maps can also be rotated. 
For example, \cite{dieleman2016exploiting} and \cite{cohen2016group} obtained rotation robustness by rotating the feature maps 3 times, by 90 degree each time, and then used average or max-pooling operations.
Besides images and feature maps, rotation robustness can also be obtained by rotating the kernel/filter.
\cite{gao2017efficient} provided an example of how to use rotated kernels,
but the weakness of their method is that the kernel size is fixed (3*3), and rotation angle is limited (45 degree each time, not arbitrary). \cite{marcos2017rotation} performed arbitrary rotations and the orientation pooling was used instead of max pooling to get the rotation-equivariance. \cite{weiler2018learning} proposed the steerable filter CNNs, which employed steerable filters to compute orientation dependent responses without suffering interpolation artifacts from filter rotation, and used group convolutions for an equivariant mapping. Recently, rotation equivariant/invariant networks have been utilized in 3D recognition tasks, such as CubeNet \citep{worrall2018cubenet}, ClusterNet \citep{chen2019clusternet}.

In contrast to these methods that aim to obtain robustness to rotations, we use the rotation equivariant/invairant networks to \nabc{negate the effects of the projection operation from the camera-view to the ground-plane.}
\nabc{Specifically, the multi-rotated filters are used to reduce the influence of the average-height projection on the extracted features, which improves the multi-view fusion counting performance.}


\section{Multi-View Counting via Multi-View Fusion}

\par For multi-view counting, we assume that the cameras are fixed, the camera calibration parameters (both intrinsic and extrinsic) are known, and that the camera frames across views are synchronized. Given the set of multi-view images, the goal is to predict a scene-level density map defined on the ground-plane of the 3D scene (see Fig. \ref{fig:abstract_pipepline}). The \abc{ground-truth} ground-plane density map is obtained in a similar way as the traditional camera-view density map -- the ground-plane annotation map is obtained using the ground-truth 3D coordinates of the people, which is then convolved by a fixed-width Gaussian kernel to obtain the density map \nabc{on the ground-plane}.

\begin{table}[t]
\centering
\begin{tabular}{ll}
\footnotesize
\begin{tabular}{|c|c|}
\hline
\multicolumn{2}{|c|}{FCN-7} \\ \hline
Layer         & Filter      \\ \hline
conv 1             & $16\! \times\! 1\! \times\!  5\!  \times\!  5$     \\ 
conv 2             & $16\!  \times\!  16\!  \times\!  5\!  \times\!  5$    \\ 
pooling   & $2\!  \times\!  2\!  $         \\ 
conv 3             & $32\!  \times\!  16\!  \times\!  5\!  \times\!  5$   \\ 
conv 4             & $32\!  \times\!  32\!  \times\!  5\!  \times\!  5$ \\ 
pooling   & $2\!  \times\!  2\!  $          \\ 
conv 5             & $64\!  \times\!  32\!  \times\!  5\!  \times\!  5$ \\ 
conv 6             & $32\!  \times\!  64\!  \times\!  5\!  \times\!  5$ \\ 
conv 7             & $1\!  \times\!  32\!  \times\!  5\!  \times\!  5$  \\ \hline
\end{tabular}
&
\footnotesize
\begin{tabular}{|c|c|}
\hline
\multicolumn{2}{|c|}{Fusion}  \\ \hline
Layer & Filter     \\ \hline
   concat   &  - \\ 
conv 1     & $64\!  \times\!  n\!  \times\!  5\!  \times\!  5$   \\ 
conv  2     & $32\!  \times\!  64\!  \times\!  5\!  \times\!  5$  \\
conv 3     & $1\!  \times\!  32\!  \times\!  5\!  \times\!  5$   \\ \hline
\end{tabular}
\end{tabular}
\caption {FCN-7 backbone and fusion module. 
The Filter dimensions are output channels, input channels, and filter size ($w\!  \times\!  h$).
}
\label{table:FCN-7_fusion_module}
\end{table}

\par
In the following two sections, we propose three fusion approaches for multi-view counting:
1) the {\em late fusion} model projects camera-view density maps onto the ground plane and then fuses them together, and requires a projection normalization step;
2) the {\em na\"ive early fusion} model projects camera-view feature maps onto the ground plane then fuses them;
3) to handle inter-view and intra-view scale variations, the {\em multi-view multi-scale early fusion} model (MVMS) selects features scales to be consistent across views when projecting to the same ground-plane point, \nabc{and uses rotation selection to handle rotation effects of the projection (MVMSR).}

\zqqblue{We note that there are differences between fusing density maps and fusing feature maps, since they contain different information. Density maps only contain location information of the crowd, but do not contain identity information of people (person appearance features). Thus, fusing density maps requires combining / aligning local density information across projected views, but may suffer from errors due to ambiguity or distortion caused by the 2D to 3D projection.
In contrast, feature maps contain identity (appearance) information that can  help to find correspondences of the same person in different views on the ground-plane. However, since the person is a different distance from each camera, this information is present at different scales among the camera views, which makes learning the correspondences more difficult because the DNN should see all combinations of scales to become scale invariant. Thus to alleviate the issue of scale variations among cameras, we propose a scale-aware fusion step, which uses  image pyramids to select features at the same scale before projection. In this way, all the features projected onto the ground-plane are at the same scale, and the relationships among features is easier to learn.}

We first present the common components, and then the 3 fusion models.

\subsection{Backbone for camera views}
\par A fully-convolutional network (denoted as FCN-7) is used on each camera view to extract image feature maps or estimate a corresponding view-level density map. The FCN-7 settings are shown in Table \ref{table:FCN-7_fusion_module}. 
For the ablation study, \zqq{CSR-Net \citep{li2018csrnet} and LCC \citep{liu2020adaptive} are also used as feature backbone (see Section \ref{text:CSRnet}).}


\subsection{Image to ground-plane projection}
\begin{figure}[t]
\begin{center}
   \includegraphics[width=\linewidth]{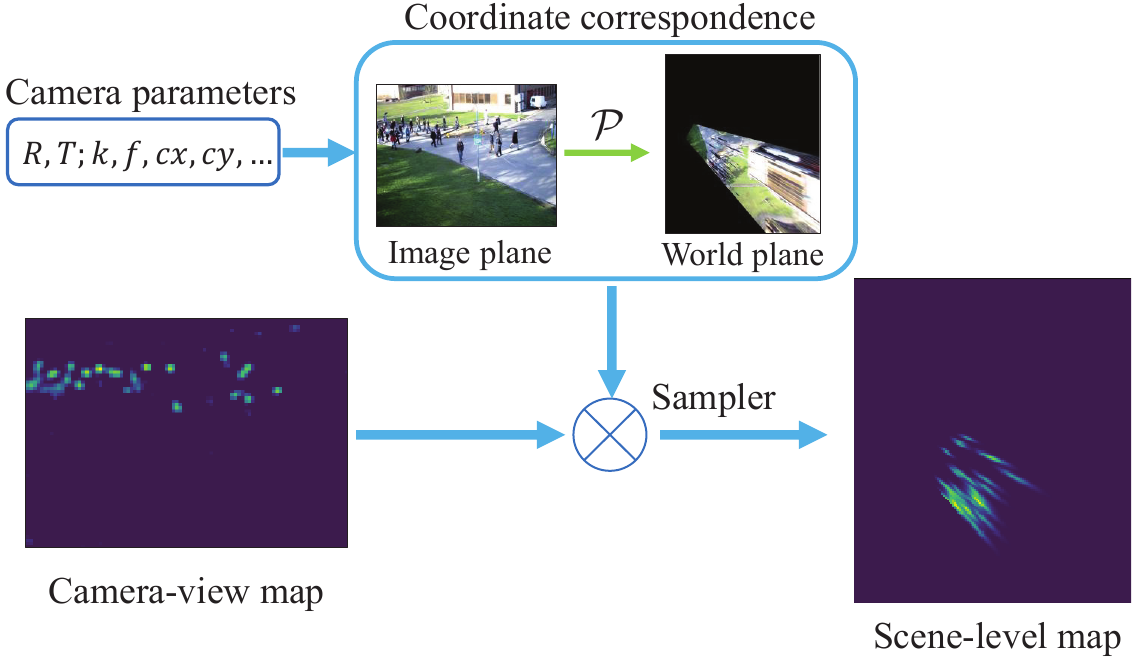}
\end{center}
   \caption{The projection module to transform camera-view maps to a ground-plane representation. Here the camera-view map is visualized as a density map.}
\label{fig:projection}
\end{figure}

\par As we assume that the intrinsic and extrinsic parameters of the cameras are known, the projection from a camera's 2D image space to a 3D ground-plane representation can be implemented as a differentiable fixed-transformation module (see Fig.~\ref{fig:projection}). The 3D height (z-coordinate) corresponding to each image pixel is unknown. Since the view-level density maps are based on head annotations and the head is typically visible even during partial occlusion, we assume that each pixel's height in the 3D world is a person's average height (1750 mm). The camera parameters together with the height assumption are used to calculate the correspondence mapping $\cal P$ between 2D image coordinates and the 3D coordinates on the 3D average-height plane. Finally, the Sampler from the Spatial Transformer Networks \citep{Jaderberg2015Spatial} is used to implement the projection, resulting in the ground-plane representation of the input map.

\begin{figure}[t]
\begin{center}
   \includegraphics[width=0.9\linewidth]{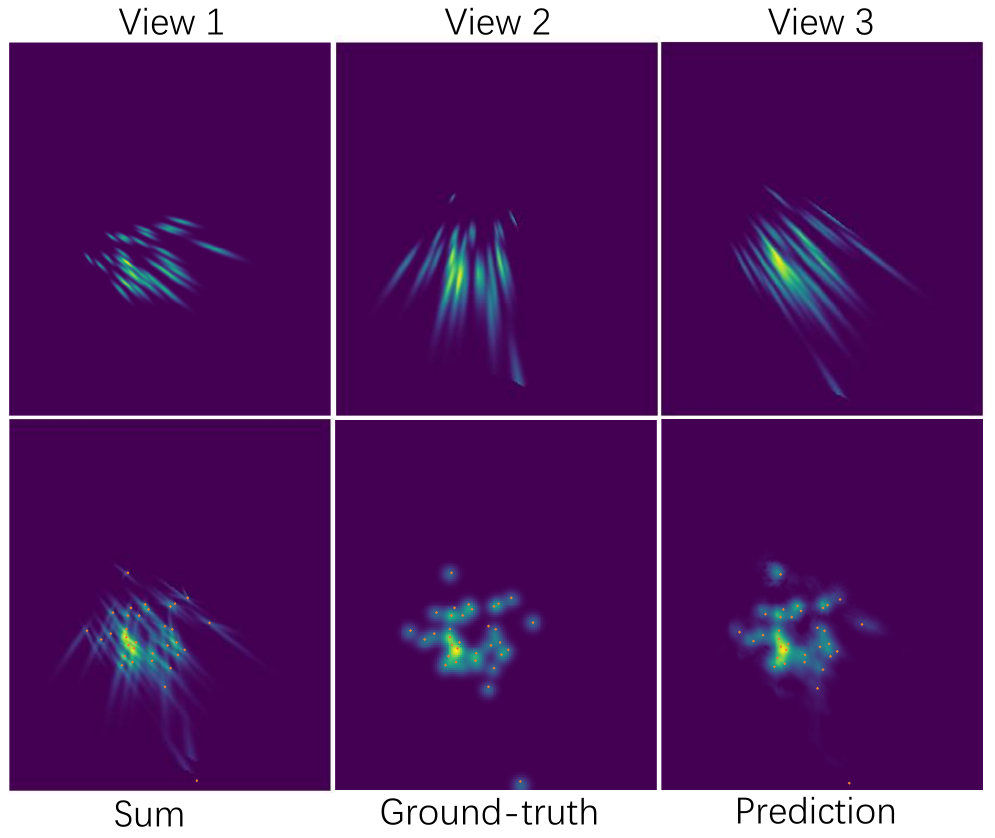}
\end{center}
   \caption{\zqqblue{Example of single-view density maps projected on to the ground-plane and their summation. The intersections of the projected Gaussians are close to the people locations on the ground-plane. The orange dots are the ground-truth annotations.}}
\label{fig:dmap_proj}
\end{figure}

\subsection{Late fusion model}
\par

and then project them to the ground-plane for fusion and obtaining the scene-level density map, where the intersections of the projected Gaussians are close to the people locations on the ground-plane (see Fig. 4) and are mapped to the Gaussian kernels of the ground-plane density map.

The main idea of the late fusion model is to first estimate the crowd density maps in each camera view,
\zqqblue{and then project them to the ground-plane for fusion and prediction of the scene-level density map. As shown in Fig.~\ref{fig:dmap_proj}, the intersections of the projected Gaussians will be close to the people locations on the ground-plane, and thus the fusion network aims to transform the intersection points into the Gaussian kernels on the ground-plane.}
In particular, the late fusion model consists of 3 stages (see Fig.~\ref{fig:two_models} top): 1) estimating the camera-view density maps using FCN-7 on each view; 2) projecting the density maps to the ground-plane representation using the projection module; 3) concatenating the projected density maps channel-wise and then applying the Fusion module to obtain the scene-level density map. The network settings for the fusion network are presented in Table \ref{table:FCN-7_fusion_module}.

\begin{figure}[t]
\begin{center}
   \includegraphics[width=\linewidth]{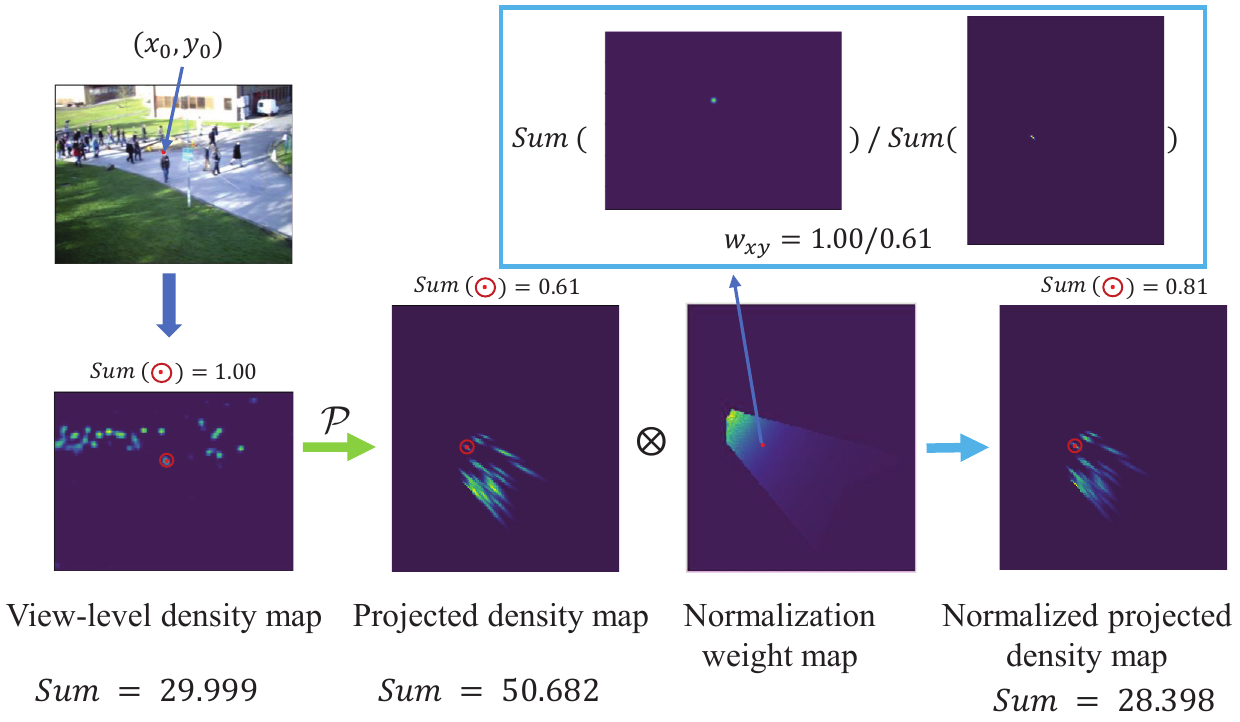}
\end{center}
   \caption{The projection normalization process for the late fusion model.
   \abc{{\em Sum} is the sum of the whole density map, while {\em Sum}(\textcolor{red}{$\odot$})} is the sum over the circled region \zqq{(diameter 5)}.
}
\label{fig:normalization}
\end{figure}

\sssection{Projection Normalization.} One problem is that the density map is stretched during the projection step, and thus the sum of the density map changes after the projection. Considering that the density map is composed of a sum of Gaussian kernels, each Gaussian is stretched differently depending on its location in the image plane. To address this problem, we propose a normalization method to ensure that the sum of each Gaussian kernel remains the same after projection (see Fig.~\ref{fig:normalization}). In particular, let $(x_0, y_0)$ and $(x, y)$ be the corresponding points in the image plane and the 3D world ground-plane representation. The normalization weight $w_{xy}$ for ground-plane position $(x, y)$ is
\begin{align}
  w_{xy} = \frac{\sum_{i,j}{D_{x_0, y_0}{(i, j)}}}{\sum_{m,n}{{\cal P}(D_{x_0, y_0}{(m, n)})}},
\end{align}
where $D_{x_0, y_0}$ denotes an image-space density map containing \textbf{only one Gaussian kernel centered} at $(x_0, y_0)$, $\cal P$ is the projection operation from image space to ground plane representation, \zqq{the summation operation is over the whole camera view map or projected ground-plane map}, and $(i, j)$ and $(m, n)$ are the image coordinates and ground-plane coordinates, respectively. The normalization map $W = [w_{xy}]$ for each camera is element-wise multiplied to the corresponding projected density map before concatenation.
As illustrated in Fig.~\ref{fig:normalization},
\zqq{to visualize the effect of the projection normalization, we choose a small circular region (diameter 5), and calculate its sum before and after  using the normalization.
After applying the projection normalization, the local ground-plane sum is more consistent with the corresponding image-based sum. Likewise, the sums of the whole ground-plane and the whole image are also more consistent after applying the normalization.}

\subsection{Na\"ive early fusion model}

\par The na\"ive early fusion model directly fuses the feature maps from all the camera-views to estimate the ground-plane density map. Similar to the late fusion model, we implement the early fusion model by replacing the density map-level fusion with feature-level fusion (see Fig.~\ref{fig:two_models} bottom). Specifically, the na\"ive early fusion model consists of 3 stages: 1) extracting feature maps from each camera view using the first 4 convolution layers of FCN-7; 2) projecting the image feature maps to the ground-plane representation using the projection module; 3) concatenating the projected feature maps and applying the Fusion module to estimate the scene-level density map. Note that the projection normalization step used in the late fusion model is not required for the early fusion model, since feature maps do not have the same interpretation of summation yielding a count.

\begin{figure*}[t]
\begin{center}
   \includegraphics[width=0.9\linewidth]{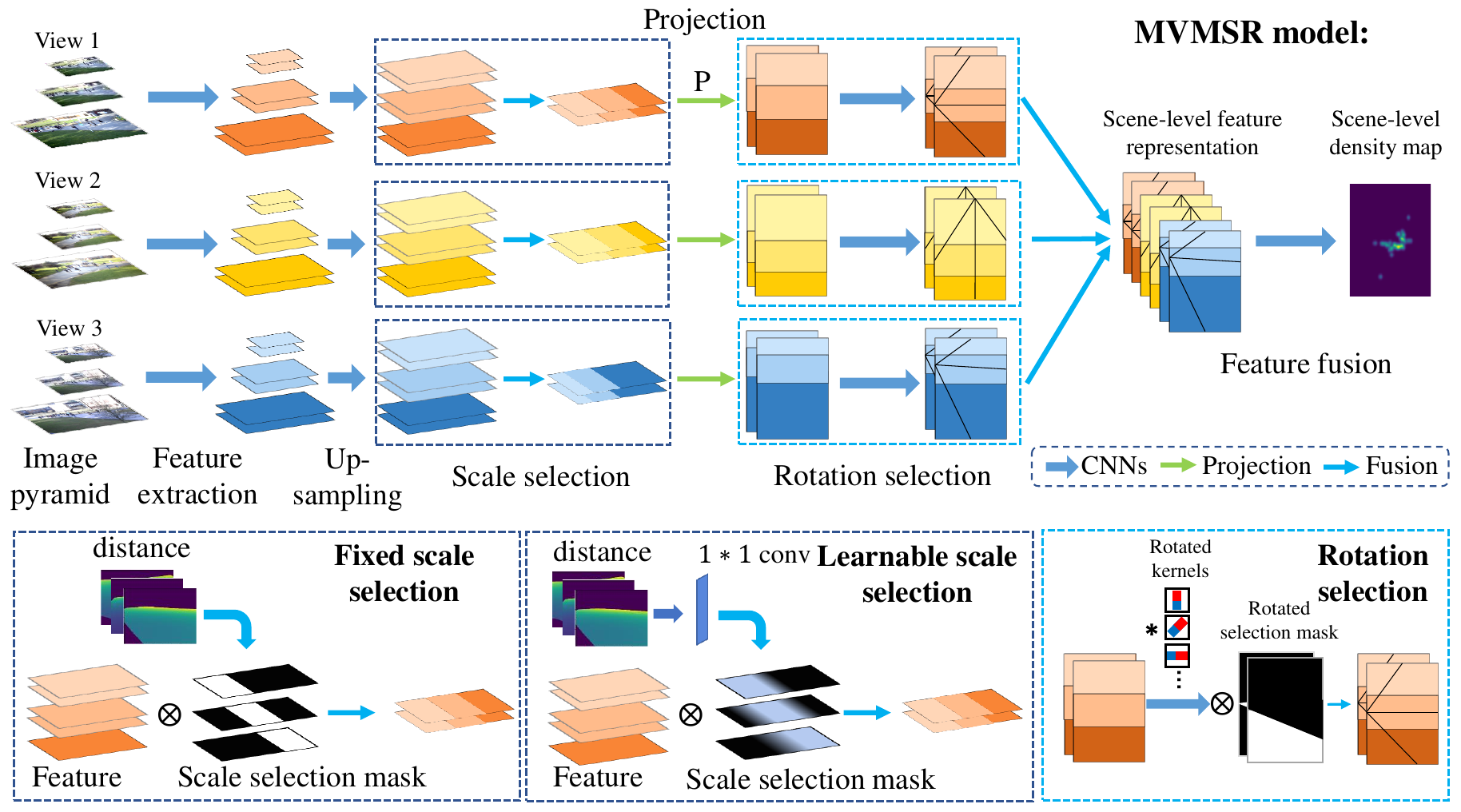}
\end{center}
   \caption{
   \add{The pipeline of multi-view multi-scale early fusion model (MVMS) with rotation selection module (MVMSR).}
   First, multi-scale feature maps are extracted with an image pyramid. The multi-scale feature maps are up-sampled to the same size. The scale selection module (the dotted box) ensures the scales of features \abc{that represent the same ground-plane point} are consistent across all views. The scale-consistent features are projected to the average-height plane and then fused to obtain the scene-level density map. Two kinds of scale selection strategies (the two dotted boxes on the right) are utilized: the fixed scale selection uses the distance information relative to a reference distance, and learnable scale selection makes the reference distance a learnable parameter. For MVMSR, a rotation selection module is added after the projection step and before the fusion step in order to remove mis-aligned rotations caused by the projection step.
   }
\label{fig:MVMS_model}
\end{figure*}
\section{Multi-view multi-scale early fusion model}
\par Intra-view scale variations are an important issue in single-view counting, as people will appear with different sizes in the image due to perspective effects. Using multiple views increases the severity of the scale variation issue; in addition to intra-view scale variation, multi-view images have inter-view scale variations, where the same person will appear at different scales across multiple views. This inter-view scale variation may cause problems during the fusion stage as there are a combinatorial number of possible scales appearing across all views, which the network needs to be invariant to. To address this problem, we extract feature maps at multiple scales, and then perform scale selection so that the projected features are at consistent scales across all views (\emph{i.e.}, a given person's features are at the same scale across all views).

\par Our proposed multi-view multi-scale (MVMS) early fusion architecture is shown in Fig.~\ref{fig:MVMS_model}. The MVMS  fusion model consists of 4 stages: 1) extracting multi-scale feature maps by applying the first 4 convolution layers of FCN-7 on an image pyramid for each camera view; 2) upsampling all the feature maps to the largest size, and then selecting the scales for each pixel in each camera-view according to the scene geometry; 3) projecting the scale-consistent feature maps to the ground-plane representation using the projection module; 4) fusing the projected features and predicting a scene-level density map using the fusion module. We consider 2 strategies for selecting the consistent scales, fixed scale selection and learnable scale selection.

\par To further boost the multi-view fusion process, a rotation selection module is added before the fusion step in the MVMS model, denoted as \emph{MVMSR}. In the rotation selection module, the projected feature maps are convolved with multiple rotated versions of the filters, and then combined by selecting among the rotated features based on the camera geometry.

\subsection{Scale selection module}
\begin{figure}[t]
\begin{center}
   \includegraphics[width=\linewidth]{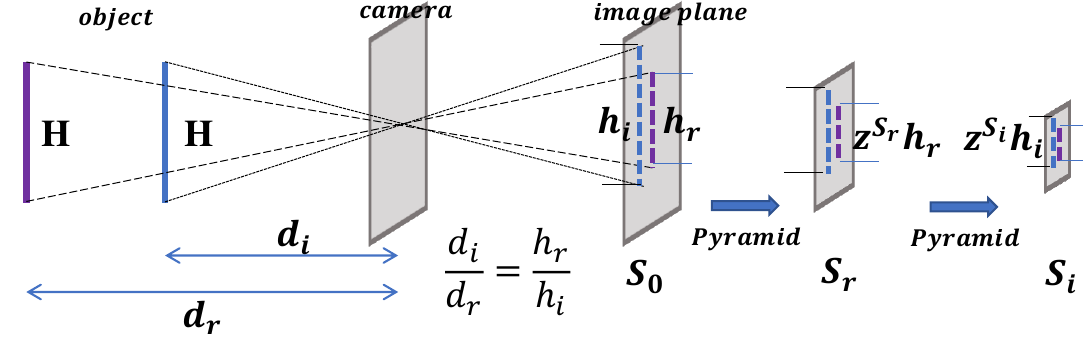}
\end{center}
   \caption{The relationship between distance-to-camera and object scale in a camera pinhole model.}
\label{fig:camera_pinhole}
\end{figure}

\begin{figure}[t!]
\begin{center}
   \includegraphics[width=\linewidth]{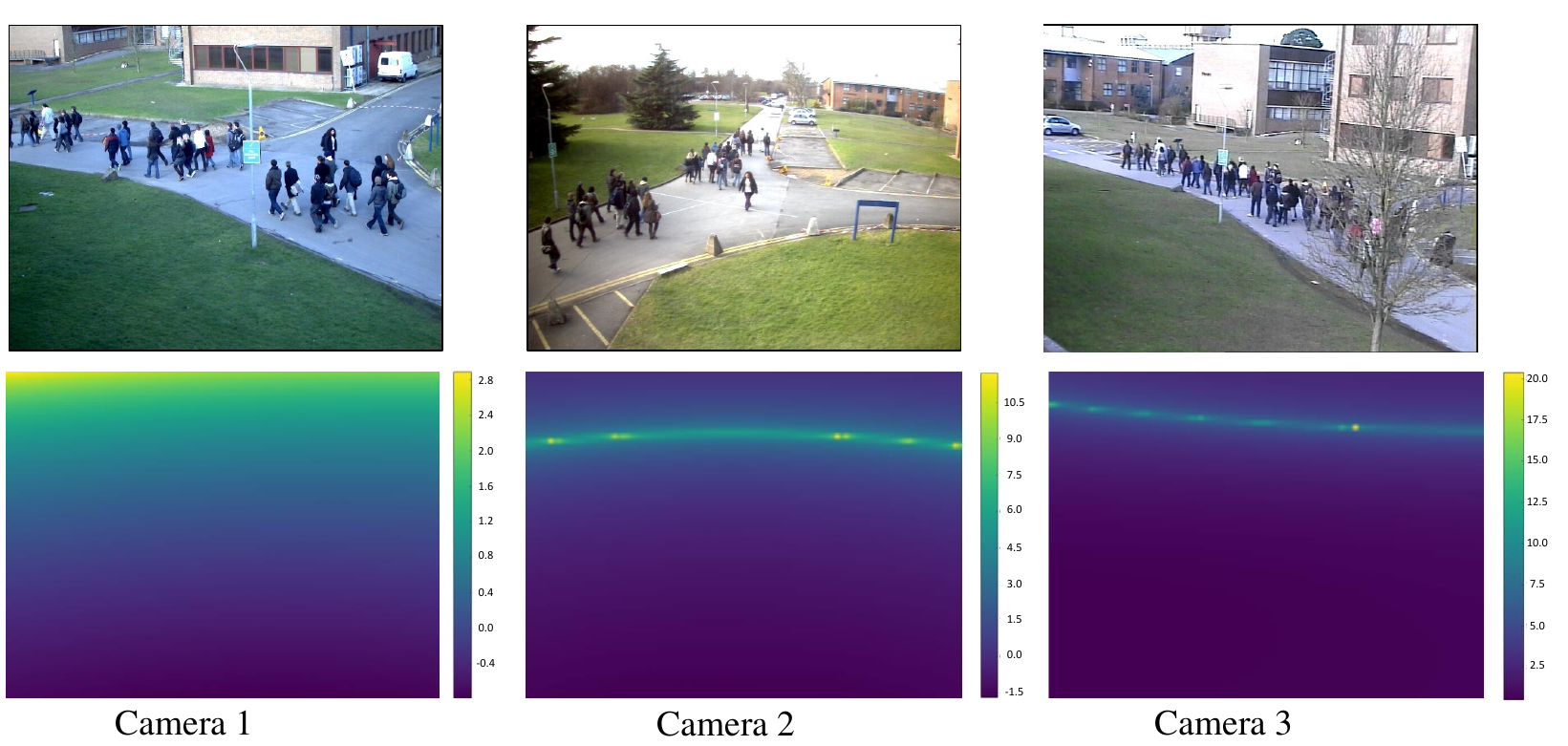}
\end{center}
   \caption{Visualization of camera distance maps of PETS2009.}
\label{fig:distance_map}
\end{figure}

In the camera pinhole model, an object's scale in an image is influenced by the object's distance to the camera (see Fig.~\ref{fig:camera_pinhole}). Therefore, the distance-to-camera information can be used to select the scale in an image pyramid to achieve scale consistency across multiple views. A distance map of each view can be calculated from the camera extrinsic parameters and the average person height. The projection operation ${\cal P}(x_0, y_0, h_{avg})$ is the projection \nnabc{of the image view coordinate $(x_0, y_0)$ to the 3D world coordinates on the average height plane $h_{avg}$}.
Then the distance-to-camera $d(x_0,y_0)$ (see Fig.~\ref{fig:distance_map}) \nnabc{is calculated by
transforming to the camera coordinate system, where the camera center is the origin,}
\begin{align}
  d(x_0, y_0) = ||R{\cal P}(x_0, y_0, h_{avg}) + T||,\label{eq:distance_map}
\end{align}
\nnabc{where $R$ and $T$ are the camera extrinsic parameters, \zq{rotation matrix and translation, respectively.}}

Next we show how to use the distance information to compute the scale according to the pinhole camera model.
Consider an image pyramid with zoom factor $z$ between neighboring scales.
Let $H$ be the height of the object in the 3D world ($H = h_{avg}$ here), and define $h_r$ as the height of the object on the image (at scale 0) when the object is at a reference distance $d_r$ from the camera.
The same object appears in the image with a different height $h_i$ when it is at distance $d_i$ from the camera.
%
%
According to camera pinhole model, we have $d_i/d_r = h_r/h_i$. In the image pyramid, the object's height is ${z^{S_r}}h_r$ in image scale $S_r$ (at distance $d_r$) and ${z^{S_i}}h_i$ in image scale $S_i$ (at distance $d_i$). Thus, to achieve scale consistency, where the heights are equal in the selected image scales, we require that
\begin{align}
{z^{S_r}}h_r = {z^{S_i}}h_i.
\end{align}
Solving for $S_i$, we obtain the scale required for the object at distance $d_i$ to be consistent with the object at reference distance $d_r$ and at reference scale $S_r$,
\begin{align}
S_i = S_r - \log_z(d_i/d_r). \label{eq:scale_sel_raw}
\end{align}

\subsubsection{Fixed scale-selection}

The fixed scale selection strategy is illustrated in Fig.~\ref{fig:MVMS_model} (bottom-left). For a given camera, let $\{F_0, \cdots , F_n\}$ be the set of feature maps extracted from the image pyramid, and then upsampled to the same size.  Here $F_0$ is the original scale and $F_n$ is the smallest scale. A distance map is computed according to (\ref{eq:distance_map}) for the camera-view, where $d(x_0,y_0)$ is the distance between the camera's 3D location and the projection of the point $(x_0, y_0)$ into the 3D-world (on the average height plane). A scale selection map $S$, where each value corresponds to the selected scale for that pixel, is computed according to (\ref{eq:scale_sel_raw}),
\begin{align}
  S(x_0, y_0) = S_{r} - \lfloor \log_z{\frac{d(x_0, y_0)}{d_r}} \rfloor,
\end{align}
where 
$\lfloor \cdot  \rfloor$ is the floor function. $d_r$ and $S_r$ are the reference distance and the corresponding reference scale number, which are the same for all camera-views. In our experiments, we set the reference distance $d_r$ as the distance value for the center pixel of the first view, and $S_r$ as the middle scale of the image pyramid. Given the scale selection map $S$, the feature maps across scales are merged into a single feature map, $F = \sum_{i} \mathbbm{1}(S=i) \otimes F_i$, where $\otimes$ is element-wise multiplication, and $\mathbbm{1}$ is an element-wise indicator function.

\begin{figure}[t!]
\begin{center}
   \includegraphics[width=\linewidth]{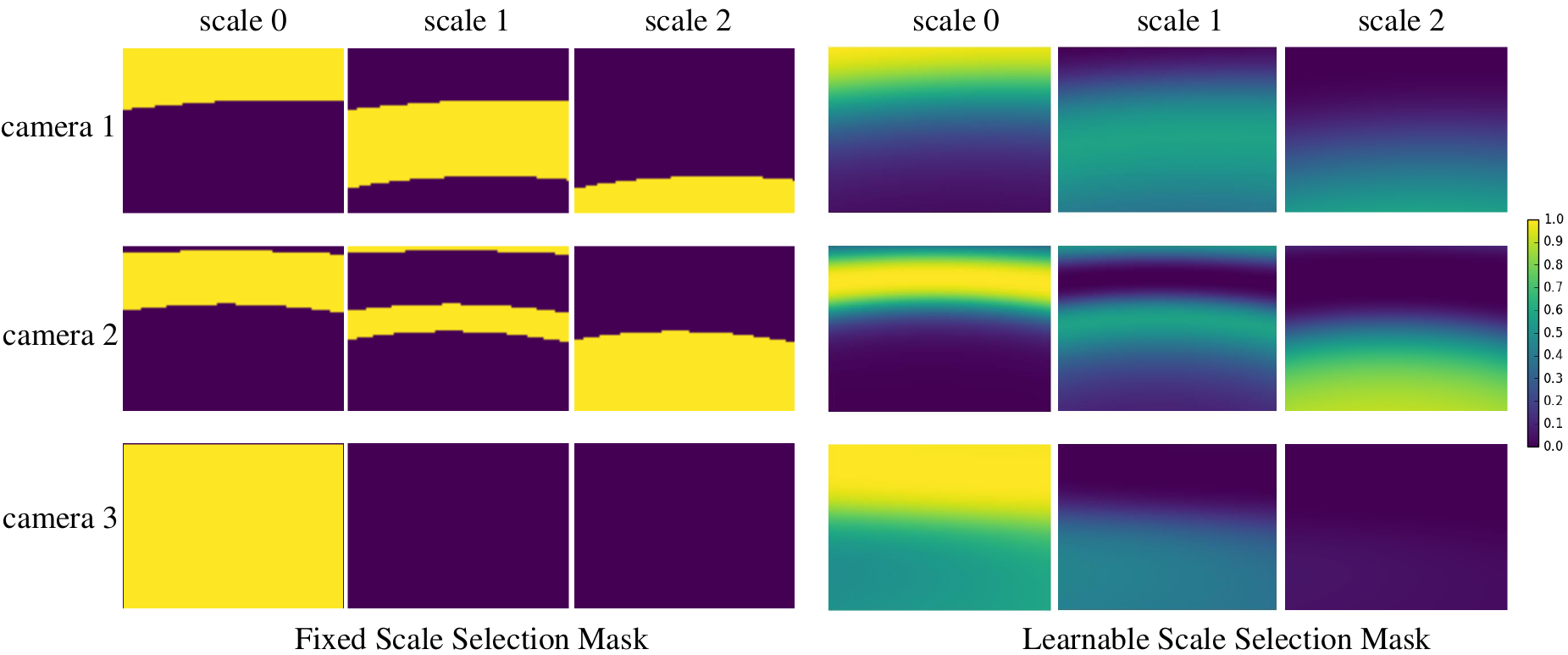}
\end{center}
   \caption{The fixed and learnable scale selection masks for PETS2009.}
\label{fig:scale_selection_mask}
\end{figure}

\subsubsection{Learnable scale-selection}

The fixed scale selection strategy requires setting the reference distance and reference scale parameters. To make the scale selection process more adaptive to the view context, a learnable scale-selection model is considered (see Fig.~\ref{fig:MVMS_model} (bottom-right)),
\begin{align}
  S(x_0, y_0) = b + k \log_z{\frac{d(x_0, y_0)}{d_r}},
\end{align}
where the learnable parameter $b$ corresponds to the reference scale, and $k$ adjusts the reference distance. The learnable scale selection can be implemented as a 1$\times$1 convolution on the log distance map. Then, a soft scale selection mask $M_i$ for scale $i$ can be obtained,
\begin{align}
  M_{i}(x_0, y_0) = \frac{e^{-(S(x_0, y_0)-i)^2}}{\sum_{j=0}^{n}e^{-(S(x_0, y_0)-j)^2}}.
\end{align}
\zqq{Note that $M_i(x_0,y_0)$ has values between 0 and 1, and sums to 1 across scales, $\sum_{i=0}^k M_i(x_0,y_0) = 1$.}
The scale consistent feature map is then
\zqq{
\begin{align}
F = \sum_{i=0}^k M_i \otimes F_i,
\end{align}
which is equivalent to a per-pixel soft-attention mechanism across scales, where $M_i$ is the pixel-wise soft-attention mask.}
%
%

\par
The fixed and learnable scale selection masks for PETS2009 are shown in Fig.~\ref{fig:scale_selection_mask}. 
The fixed scale selection produces binary masks and learnable scale selection produces soft masks. The learnable scale selection gives more freedom for the networks to fuse the multi-scale features, especially on the edges of each scale's masks.

\begin{figure}[t]
\begin{center}
   \includegraphics[width=\linewidth]{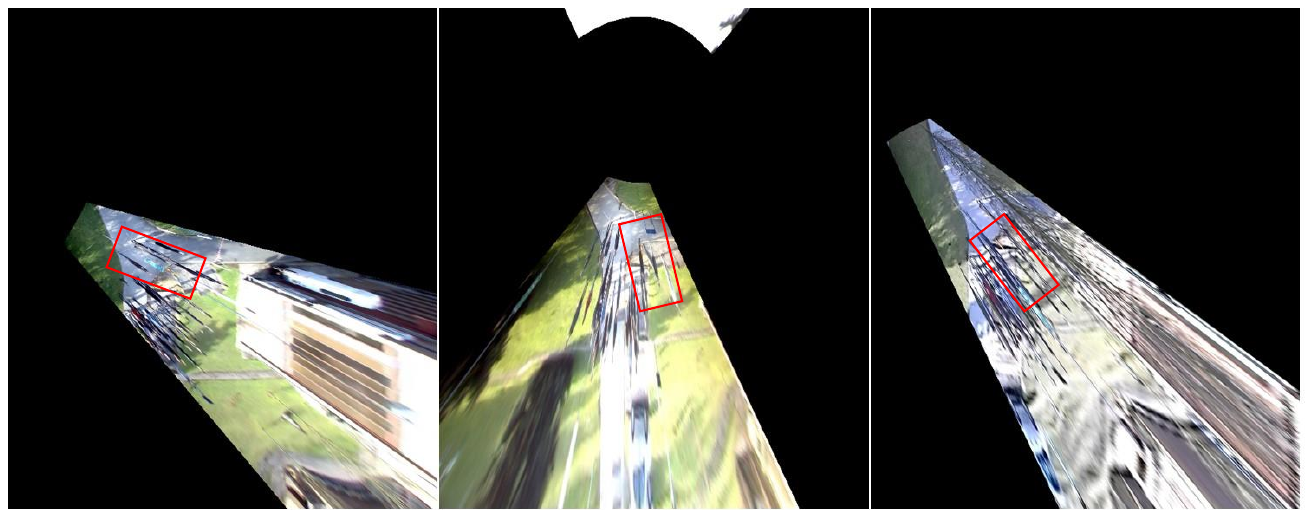}
\end{center}
   \caption{Rotation distortion of projected features caused by the average-height projection (using the original image for visualization).}
\vspace{-0.2cm}
\label{fig:rotation_effect}
\end{figure}
\subsection{Rotation selection module}

\begin{figure}[t]
\begin{center}
   \includegraphics[width=\linewidth]{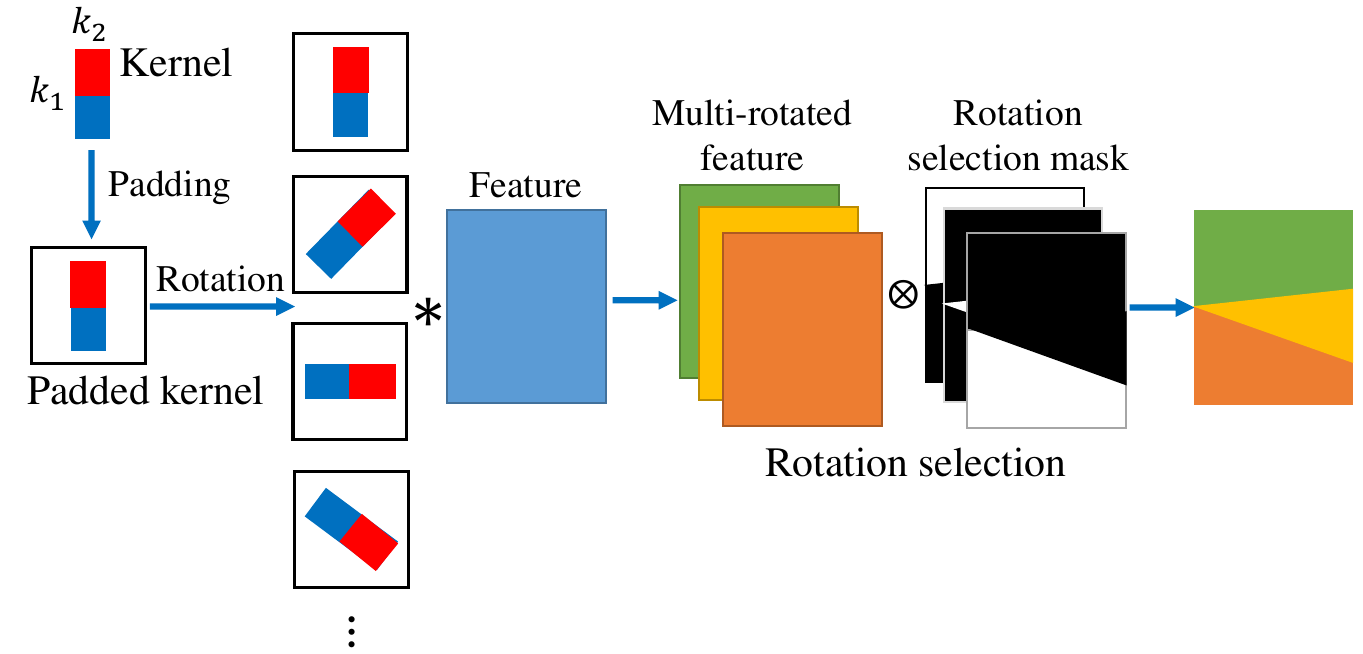}
\end{center}
   \caption{The rotation selection layer. The same kernel is padded, rotated and then convolved with the projected features. The rotation selection mask is used to select and fuse the multi-rotated features.}
\vspace{-0.6cm}
\label{fig:rotation_selection_layer}
\end{figure}

\begin{figure}[t]
\begin{center}
   \includegraphics[width=0.8\linewidth]{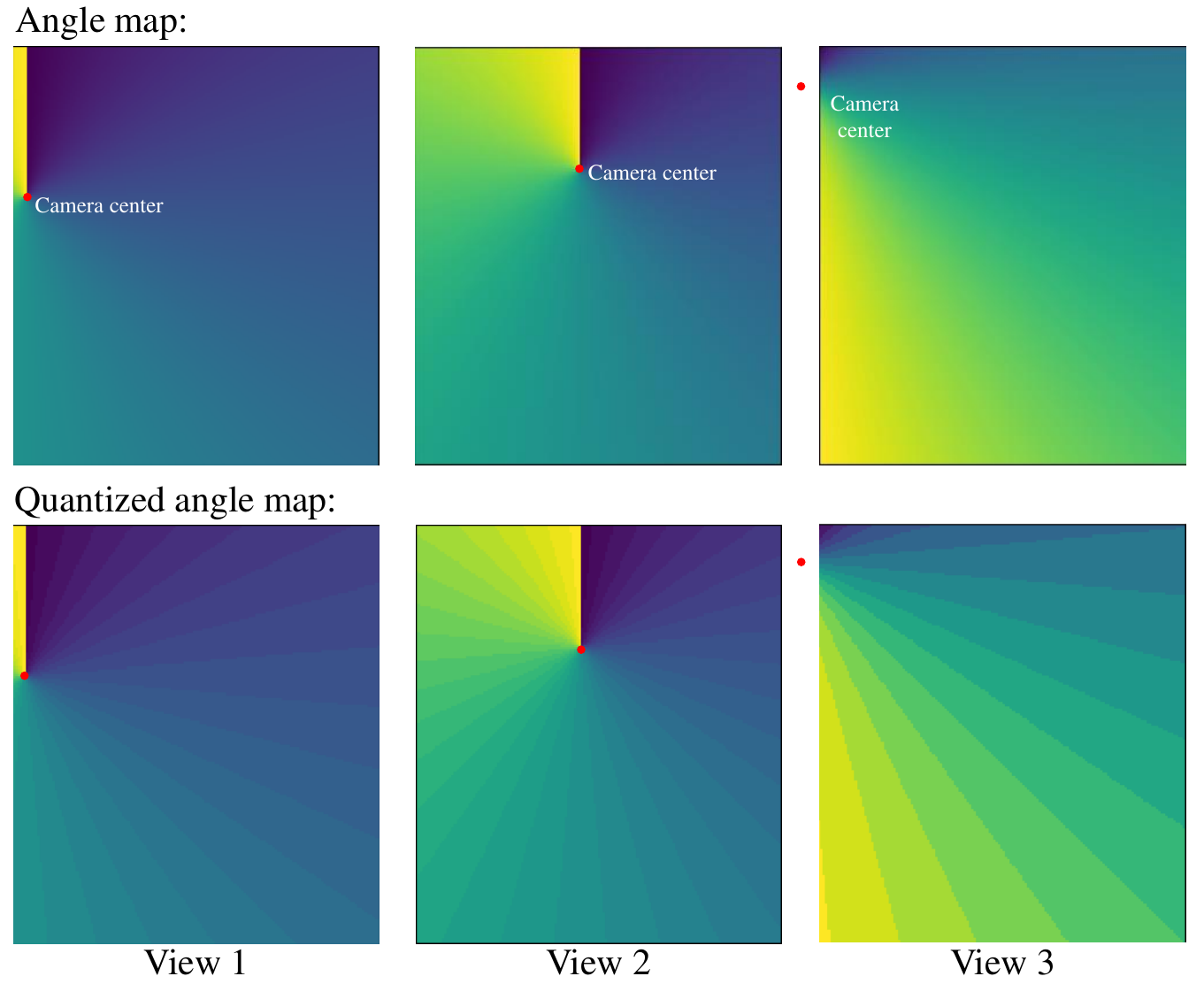}
\end{center}
\vspace{-0.2cm}
   \caption{Examples of the rotation selection masks for PETS2009. The first row shows the angle maps for each view, which show the rotation angle for each pixel. Brighter color means larger rotation angles. The second row shows the quantized angle maps, where a finite number of rotation angles are used to reduce computation complexity.}
\vspace{-0.6cm}
\label{fig:rotation_selection_mask}
\end{figure}

\par
In the projection module, the average-height assumption is utilized in which all pixels in each view are assumed to have the average height. The average-height projection is applicable due to the head annotations in the datasets. On the other hand, the average height projection makes the feature patterns stretched along the view ray. Therefore, the features are ``rotated'' to the view ray direction after the projection (see Fig.~\ref{fig:rotation_effect}). To further improve the multi-view fusion, a rotation selection module is proposed and used before the multi-view fusion step in order to counteract this phenomenon.

The rotation selection layer is illustrated in Fig. \ref{fig:rotation_selection_layer}, where ``tall'' aspect-ratio kernels ($k_1$ by $k_2$) are adopted due to the stretched patterns. First, the kernel is padded to be square to ensure the feasibility of the arbitrary rotation of the kernel, where the square size is $\lceil \sqrt{2}\max(k_1, k_2) \rceil$. Second, the kernels are rotated over a range of angles $\{r_0, r_1, \cdots, r_m\}$ (decided by the rotation selection map, see next paragraph), which can be implemented using the Sampler from \citep{Jaderberg2015Spatial}. Third, each rotated kernel is convolved with the projected  feature maps, resulting in multi-rotated features $\{F_0, ..., F_m\}$. Finally, the multi-rotated features are selected 
and fused with rotation selection masks.

The rotation selection mask is calculated with the camera parameters on the \zq{average-height} plane. 
Suppose $W_c = (x_c, y_c)$ are the coordinates on the scene-level plane (after projection), and the $(R,T)$ are the camera extrinsic parameters. Therefore, the camera location is $O = -{R^T}T$, and the corresponding view ray is along $\overrightarrow{OW_c}$. The rotation angle $r(x_c, y_c)$ is the angle between the unit vector $(0, 1)$ along $y$ direction and $\overrightarrow{OW_c}$. After the rotation angle map $r(x_c, y_c)$ is calculated, it is then quantized by $q$ degree into the rotation range $\{r_0, r_1, ..., r_m\}$. The rotation selection mask for rotation angle $r_i$ is $\mathbbm{1}(r=r_i)$, and the fused features are $F = \sum_{i} \mathbbm{1}(r=r_i) \otimes F_i$. Examples of the rotation selection masks are presented in the second row of Fig.~\ref{fig:rotation_selection_mask}.

\subsection{Training details}

\begin{figure*}[t]
\begin{center}
   \includegraphics[width=0.85\linewidth]{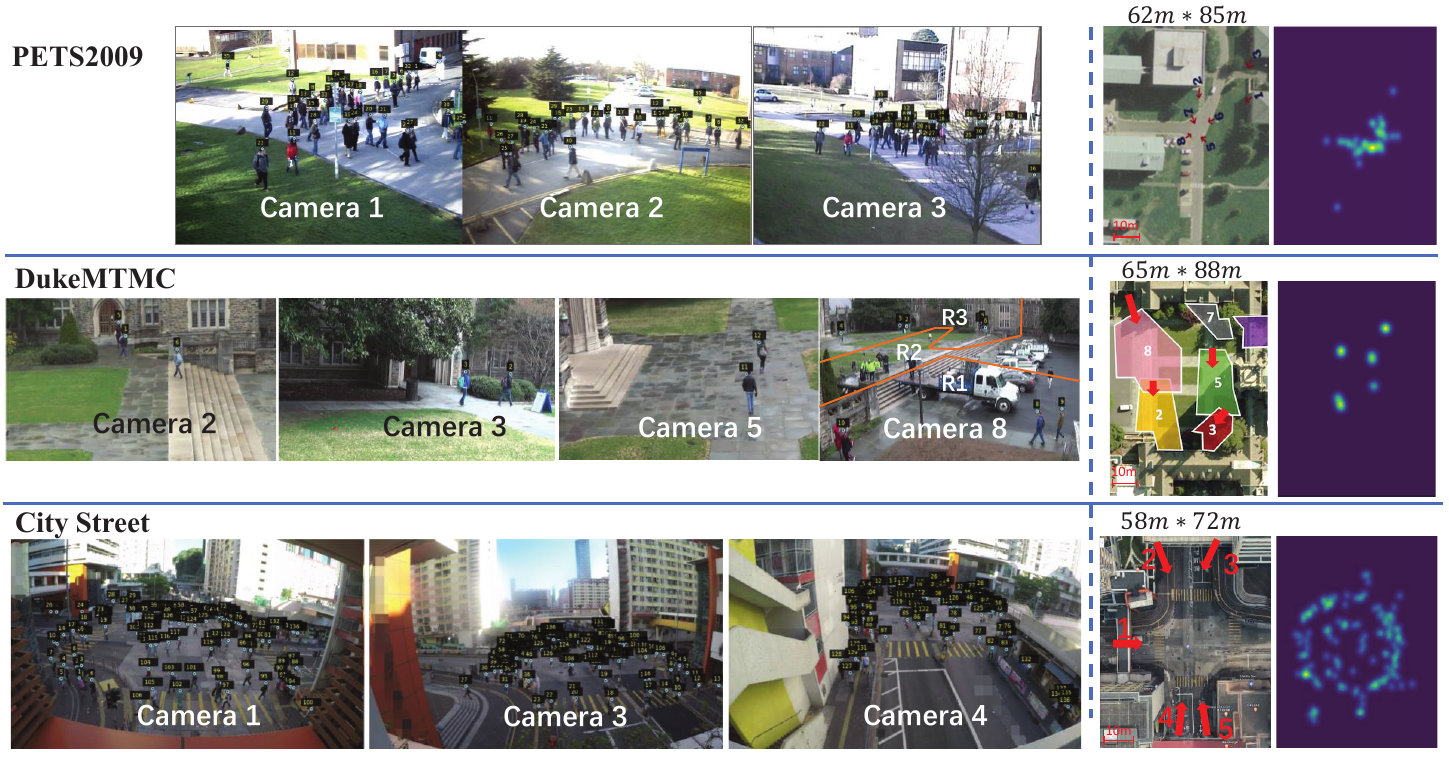}
\end{center}
   \caption{Examples from 3 multi-view counting datasets. The first column shows the camera frames and annotations. The second column shows the camera layout and scene-level ground-plane density maps.
   \zqq{Note that `R2' region of Camera 8 of DukeMTMC dataset is not used since no annotations are available.}
   }
\vspace{-0.2cm}
\label{fig:datasets}
\end{figure*}

\par
A two-stage procedure is used to train the model. The first stage trains the main scene-level density map estimation task as well as auxiliary view-level density map estimation tasks.
The auxiliary task for late fusion consists of auxiliary losses applied between the predicted and GT density maps for each view.
The auxiliary task for early fusion uses an auxiliary branch of 3 layers of FCN to predict density maps for each view, followed by an auxiliary loss on the view-level predicted density maps. The learning rate is set to 1e-4. In the second stage, the auxiliary view-level density map estimation tasks are removed, leaving only the scene-level task. FCN-7 (either density map estimator or feature extractor) is fixed and the fusion and scale selection parts are trained. The loss function is the pixel-wise squared error between the ground-truth and predicted density maps. The learning rate is set to 1e-4, and decreases to 5e-5 during training. \abc{After training the two stages, the model is fine-tuned end-to-end.}
The training batch-size is set to 1 in all experiments.


\begin{table}
\centering
\begin{tabular}{@{}l@{\hspace{0.2cm}}c@{\hspace{0.2cm}}c@{\hspace{0.2cm}}c@{\hspace{0.2cm}}c@{}}
\hline
 Dataset  & resolution  & view & train / test & crowd \\
\hline
 PETS2009  & $768\!\times\!576$ & 3  & 1105 / 794 & 20-40 \\  
 DukeMTMC  &$1920\!\times\!1080$  & 4  & 700 / 289   & 10-30 \\ 
 CityStreet   &$2704\!\times\!1520$  & 3 & 300 / 200 & 70-150 \\
\hline
 \end{tabular}
 \caption{The comparison of three multi-view datasets.}
\label{table:datasets}
\vspace{-0.8cm}
\end{table}

\section{Datasets and Experiment Setup}
In this section we introduce the 3 multi-view counting datasets and the experiment settings.

\subsection{Datasets}
We test the proposed multi-view counting framework on two existing datasets, PETS2009 and DukeMTMC, and our newly collected CityStreet dataset. Table \ref{table:datasets} provides a summary, and Fig. \ref{fig:datasets} shows examples.

\subsubsection{PETS2009}
PETS2009 \citep{ferryman2009pets2009} is a multi-view sequence dataset containing crowd activities from 8 views. The first 3 views are used for the experiments, as the other 5 views have low camera angle, poor image quality, or unstable frame rate. To balance the crowd levels, we use sequences S1L3 (14\_17, 14\_33), S2L2 (14\_55) and S2L3 (14\_41) for training (1105 images in total), and S1L1 (13\_57, 13\_59), S1L2 (14\_06, 14\_31) for testing (794 images). The calibration parameters (extrinsic and intrinsic) for the cameras are provided with the dataset. To obtain the annotations across all views, we use the View 1 annotations provided by \cite{leal2015motchallenge} and project them to other views, followed by manual annotations to get all the people heads in the images.

\subsubsection{DukeMTMC}
DukeMTMC \citep{ristani2016MTMC} is a multi-view video dataset for multi-view tracking, human detection, and ReID. The multi-view video dataset has videos from 8 synchronized cameras for 85 minutes with 1080p resolution at 60 fps. For our counting experiments, we use 4 cameras that have overlapping fields-of-view (cameras 2, 3, 5 and 8). The synchronized videos are sampled every 3 seconds, resulting in 989 multi-view images. The first 700 images are used for training and the remaining 289 for testing.
\zq{Using the same sampling method for creating the DukeMTMC training set, 200 multi-view frames are extracted from DukeMTMC Test Hard set for extra evaluation.}
Camera extrinsic and homography parameters are provided by the dataset. In the original dataset, annotations for each view are only provided in the view ROIs, which are all non-overlapping on the ground-plane and \zqq{in Camera 8, only R3 region is used and R1 and R2 are excluded}. Since we are interested in overlapping cameras, we project the annotations from each camera view to the overlapping areas in all other views. Region R2 (see Fig. \ref{fig:datasets}) is excluded during the experiment, since there are no annotations provided there.

\subsubsection{CityStreet}
We collected a multi-view video dataset of a busy city street in Hong Kong using 5 synchronized cameras.  The videos are about 1 hour long with 2.7k ($2704\!\times\!1520$) resolution at 30 fps. We select Cameras 1, 3 and 4 for the experiment (see Fig.~\ref{fig:datasets} bottom).
\abcn{The cameras' intrinsic and extrinsic parameters are estimated using the calibration algorithm from \cite{zhang2000flexible}}.
500 multi-view images are uniformly sampled from the videos, and the first 300 are used for training and remaining 200 for testing.
The ground-truth 2D and 3D annotations are obtained as follows.
\zq{The head positions of the first camera-view are annotated manually, and then projected to other views and adjusted manually.
Next, for the second camera view, new people (not seen in the first view), are also annotated and then projected to the other views.
This process is repeated until all people in the scene are annotated and associated across all camera views.}
Our dataset has larger crowd numbers (70-150), compared with PETS (20-40) and DukeMTMC (10-30). 
Our dataset also contains more crowd scale variations and occlusions due to vehicles and fixed structures.



\subsection{Experiment setup}

\subsubsection{Ground-truth settings}

The ground-truth scene-level density maps are created \zqq{by convolving the people's ground-plane annotation map with  a fixed-bandwidth Gaussian kernel. The people's ground-plane annotations are estimated from their camera-view annotations and the camera calibration information. First, each person's height in 3D is estimated by
selecting a height, among a candidate set in the range [1.6m, 2.0m] (step of 0.1m),
so as to minimize the differences among the ground-plane coordinates obtained by projecting that person's 2D camera-view annotations to the 3D world. Second, with the estimated person's height, a person's ground-plane annotation is set to the average of that person's 2D annotations projected to the 3D world.
To construct the ground-truth scene-level density maps, we follow single-image counting and choose a fixed-bandwidth $\sigma=3$ for the Gaussian kernel.
}

The image resolutions ($w\! \times\! h$) used in the experiments are: $384\!\times\!288$ for PETS2009, $640\!\times\!360$ for DukeMTMC, and $676\!\times\!380$ for CityStreet. The resolutions of the scene-level ground-plane density maps are: $152\!\times\!177$ for PETS2009, $160\!\times\!120$ for DukeMTMC and $160\times192$ for CityStreet. For the detection baseline, the original image resolutions are used (Faster-RCNN will resize the images).

\begin{table*}
\centering
\begin{tabular}{ll|c|c|c|c|c}
    \hline
   Dataset & Method	         & MSE	& NAE & MAE/GAME(0) & GAME(1) & GAME(2)\\
   \hline

   \multirow{8}{*}{PETS2009}
   & Dmap weighted	         & 7.29	& 0.182 & 5.62	& - & -	\\
   & Detection+ReID	         & 7.01	& 0.174 & 5.46	& - & -	\\

   & Feature concatenation	 & 8.96	    & 0.300 & 7.32	& - & -	\\
   & Stitching	             & 14.51    & 0.337 & 10.90	& - & -	\\

   &Late fusion              & 5.18	& 0.138 & 3.92	& 6.38 & 7.75	\\
   &Na\"ive early fusion     & 6.76	& 0.199 & 5.42	& 7.26 & 8.13	\\
   &MVMS             & \textbf{4.83}	& \textbf{0.124} & \textbf{3.49}	& \textbf{5.30} &\textbf{ 6.27}	\\
   &MVMSR           & 4.93	& 0.130 & 3.62	& 5.37 & 6.98	\\

   \hline
   \multirow{8}{*}{DukeMTMC}
   &Dmap weighted	          & 2.06 & 0.186 & 1.54	& - & -	\\
   & Detection+ReID	          & 2.30	& 0.355 & 1.89	& - & -	\\

   & Feature concatenation	  & 5.23	& 1.030 & 4.44	& - & -	\\
   & Stitching	             & 1.65    & 0.215 & 1.24	& - & -	\\

   &Late fusion               &  1.63	& 0.187  & 1.29	& 1.77 & 2.22	\\
   &Na\"ive early fusion      & 1.90	    & 0.199 & 1.47	& 2.00 & 2.62	\\
   &MVMS	                  & 1.28   	    & 0.122 & 0.95	& 1.24 & 1.50	\\
   &MVMSR                     & \textbf{1.17}	    & \textbf{0.118} & \textbf{0.89}	& \textbf{1.19} & \textbf{1.42}	\\

   \hline
   \multirow{8}{*}{CityStreet}
   & Dmap weighted	          & 11.46	& 0.120 & 9.36	& - & -	\\
   & Detection+ReID	          & 21.18	& 0.193 & 17.48	& - & -	\\

   & Feature concatenation	 & 21.34	& 0.245 & 18.33	& - & -	\\
   & Stitching	             & 10.55    & 0.107 & 8.76	& - & -	\\


   &Late fusion              & 9.63	& 0.099 & 8.06	& 12.75 & 23.10	\\
   &Na\"ive early fusion     & 9.85	& 0.100 & 8.11	& 12.73 & 22.93	\\
   & MVMS	                  & 9.02	& 0.096 & 7.36	& 11.95 & 20.44	\\
   & MVMSR                    & \textbf{8.49}	& \textbf{0.086} & \textbf{6.98}	& \textbf{11.39} & \textbf{19.79}	\\
   \hline
\end{tabular}
\caption{ \zqq{The scene-level counting performance of different methods on the 3 datasets. MSE, NAE, MAE and GAME are used as evaluation metrics. Note that for comparison methods whose outputs are not density maps, GAME(1) and GAME(2) are not applicable. FCN-7 is used as feature backbone for PETS2009, and CSR-net is used as feature backbone for CityStreet and DukeMTMC. For MVMSR, the filter number is 32, the layer number is 3 and the quantization angle is $10^{\circ}$, $45^{\circ}$ and $45^{\circ}$ for the PETS2009, DukeMTMC and CityStreet, respectively.
}}
\label{tab:scene_count}
\end{table*}

\subsubsection{Methods}
We test our multi-view fusion models, denoted as ``Late fusion'', ``Na\"ive early fusion'',  ``MVMS'' (multi-view multi-scale early fusion), \nabc{and ``MVMSR'' (MVMS with rotation selection)}. The late fusion model uses projection normalization. MVMS uses learnable scale selection, and a 3-scale image pyramid with zoom factor of 0.5. \add{MVMSR uses 3 rotation selection layers with filter number $F=32$ and quantization angle $Q = 10^{\circ}$, $45^\circ$ and $45^\circ$ for PETS2009, DukeMTMC and CityStreet, respectively}. Besides, we have also performed the models with different feature extraction backbones. These settings will be tested later in the ablation study.

\par For comparisons, we test and compare with several comparison methods.
The first comparison method is an approach to fusing camera-view density maps into a scene-level crowd count, denoted as ``Dmap weighted'', which is an adaptation from \cite{Ryan2014Scene}. First single image counting model is applied to get the density map $D_i$ for each camera-view. The density maps are then fused into a scene-level count using a weight map $W_i$ for each view,
	\begin{align}
	C = \sum_i \sum_{x_0,y_0} W_i(x_0,y_0) D_i(x_0,y_0),
	\end{align}
where the summations are over the camera-views and the image pixels. The weight map $W_i$ is constructed based on how many views can see a particular pixel. In other words,  $W_i(x_0, y_0) = 1 / t$, where $t$ is the number of views that can see the projected point ${\cal P}(x_0,y_0)$. Note that \cite{Ryan2014Scene} used this simple fusion approach with traditional regression-based counting (in their setting, the $D_i$ map is based on the \abc{predicted counts for crowd blobs}). We also test on the methods with different single-image counting models.
Here, we are using recent DNN-based methods (CSR-net) and crowd density maps, which outperform traditional regression-based counting, and hence form a stronger baseline method compared to \cite{Ryan2014Scene}.

The second comparison method is using human detection methods and person re-identification (ReID), denoted as ``Detection + ReID''. First, Faster-RCNN \citep{ren2015faster} is used to detect humans in each camera-view. Next, the scene geometry constraints and the ReID method
Circle2020 (\citep{sun2020circle,wang2018learning}) are used to associate the same people across views. Specifically, each detection box's top-center point in one view is projected to other views, and ReID is performed between the original detection box and detection boxes near the projected point in other views. Finally, the scene-level people count is obtained by counting the number of unique people among the detection boxes in all views. The bounding boxes needed for training are created with the head annotations and the perspective map of each view.

\zqq{The third comparison method is to simply concatenate the features of the different views and directly regress a scene-level count. We've used the LCC \citep{liu2020adaptive} as the feature extractor, where the output of the scale-aware module of LCC is used as extracted features.}

\zqq{The fourth comparison method is to ``stitch'' together the counts from different camera views. In particular, first a set of non-overlapping ROIs on the
ground-plane are formed by assigning ground-plane pixels to the closest camera. These ROIs are then projected into their camera views. Next, single-view counting (CSR-net) is performed on each camera-view, and the ROI count in each camera view is obtained. Finally, the scene-level count is the sum of the ROI counts.}

\subsubsection{Evaluation}
\zqq{
The mean absolute error (MAE), mean squared error (MSE), and normalized (relative mean) absolute error (NAE) are used to evaluate multi-view counting performance, comparing the scene-level predicted counts and the ground-truth scene-level counts. Besides, Grid
Average Mean absolute Error (GAME) \citep{guerrero2015extremely} is also used to evaluate the local counting performance of the predicted scene-level density maps of the proposed methods. The definitions of these evaluation metrics are as follows.
\begin{align}
    MAE = \frac{1}{N} \sum_{i}^{N} |c_i - {\hat{c}}_{i}|,
\end{align}
\begin{align}
    MSE = \sqrt{\frac{1}{N} \sum_{i}^{N} (c_i - {\hat{c}}_{i})^2},
\end{align}
\begin{align}
    NAE = \frac{1}{N} \sum_{i}^{N} |c_i - {\hat{c}}_{i}|/{\hat{c}}_{i},
\end{align}
\begin{align}
    GAME(L) = \frac{1}{N} \sum_{i}^{N} (\sum_{l=1}^{4^L} |c_i^l - {\hat{c}}_{i}^l|),
\end{align}
where $N$ is the number of the test images, $c_i$ and $\hat{c}_{i}$ are the estimated and ground-truth people count in the $i$-th image.
As to GAME metric, the scene-level density maps are divided in $4^L$ non-overlapping patches and compute the average of the $MAE$ of these patches.
$c_i^l$ and ${\hat{c}}_{i}^l$ are the estimated and ground-truth people count of the patch $l$ of $i$-th image.
Note that $MAE$ equals the $GAME$ when $L=0$.}

In addition, we also evaluate the predicted counts in each camera-view. The ground-truth count for each camera-view is obtained by summing the ground-truth scene-level density map over the region covered by the camera's field-of-view. Note that people that are totally occluded from the camera, but still within its field-of-view, are still counted.

\section{Experiment Results}

In this section, the scene-level counting performance of the proposed DNN-based multi-view fusion methods are evaluated against other multi-camera counting methods. We also demonstrate the single-view counting performance using multi-view cameras. 
In terms of both evaluation perspectives, the proposed method can achieve better counting results on all 3 multi-view counting datasets.

\begin{table}[t]
\centering
\begin{tabular}{lccc}
\hline
 Method  &  MSE &  NAE &  MAE\\
 \hline
  Hybrid \citep{dittrich2017people} &- & -&  2.03 \\
  Late fusion (w/ PN) &2.56  &0.241 & 1.53 \\
  Naive early fusion &2.14 &0.203 &   1.71 \\
  MVMS  &1.26& 0.091 &    0.98  \\
  MVMSR  &\textbf{ 1.24} & \textbf{0.089} &    \textbf{0.96}   \\

\hline
\end{tabular}

\caption{Extra experiment results on PETS S1L1 (views 1 and 2) comparing with a traditional multi-view method. `PN' means projection normalization.}
\label{table:comprison_traditional}
\vspace{-0.5cm}
\end{table}

\subsection{Scene-level counting performance}
\begin{figure*}[t]
\begin{center}
   \includegraphics[width=0.7\linewidth]{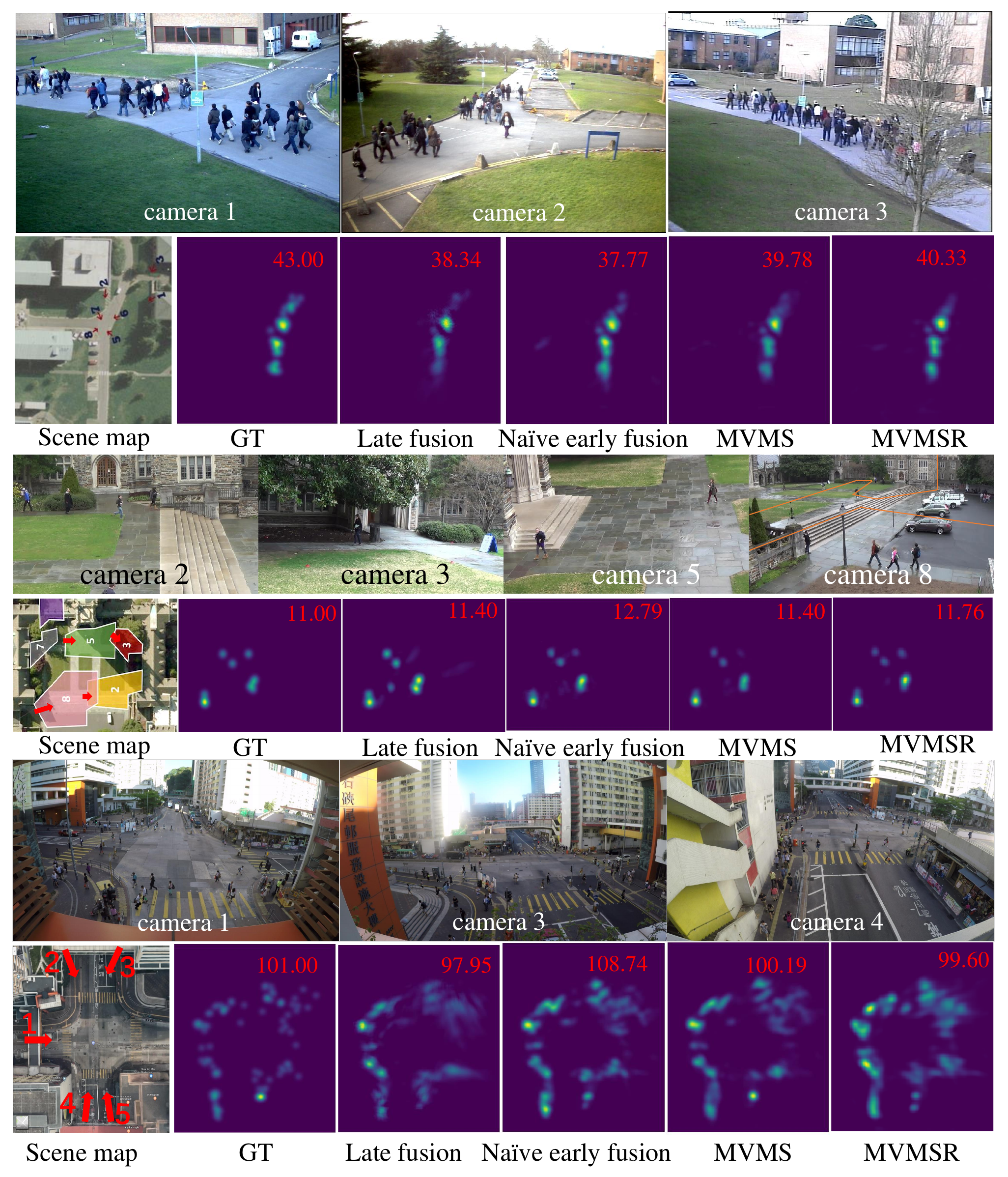}
\end{center}
\caption{The visualization results of the proposed multi-view fusion counting methods on PETS2009, DukeMTMC and CityStreet.}
\label{fig:results_visualization}
\end{figure*}

In this section, we test the proposed multi-view counting models in terms of scene-level counting performance on the 3 multi-view counting datasets, PETS2009 \citep{ferryman2009pets2009}, DukeMTMC \citep{ristani2016MTMC} and CityStreet.
The results are presented in Tables \ref{tab:scene_count}, \ref{table:PETS2009_results}, \ref{table:DukeMTMC_results}, and  \ref{table:CityStreet_results}, and examples shown in Fig.~\ref{fig:results_visualization}.


\begin{table}[t]
\centering
\begin{tabular}{lccc}
\hline
    Method                    &  MSE &  NAE &  MAE\\
    \hline
  Dmap weighted               & 7.21 &0.437 & 4.17 \\
   Detection+ReID             & 7.06 &0.371 & 3.71 \\
  Late fusion (w/ PN)         & \textbf{4.82} & \textbf{0.307} & \textbf{2.62} \\
  Na\"ive early fusion            & 6.13  &  0.361 &   2.82 \\
  MVMS   &  6.20 &0.328   &2.81 \\
  MVMSR  &6.00 &0.329 &2.89 \\

\hline
\end{tabular}
\caption{Extra experiment results on DukeMTMC Test Hard set.}
\label{table:DukeMTMC_Test_Hard_set}
\vspace{-0.5cm}
\end{table}

\subsubsection{PETS2009}
\par 
The scene-level counting results on PETS2009 are shown in Table \ref{tab:scene_count} (top row) and Table \ref{table:PETS2009_results} (``Scene'' column). On PETS2009, our proposed multi-view fusion models (use FCN-7 as backbone) achieve better results than the two comparison methods. Detection+ReID (Circle2020) performs worst on this dataset because the people are close together in a crowd, and occlusion causes severe misdetection. Among our three multi-view fusion models, na\"ive early fusion performs worse, which suggests that the scale variations in multi-view images limits the performance. Furthermore, MVMS performs much better than other models, which shows the multi-scale framework with scale selection strategies can improve the feature-level fusion to achieve better performance.

\par The performance of using one camera for the scene-level counting task is not as good as using multi-cameras.
In particular, using Dmap with cameras 1 or 2 performs poorly due to the limited field-of-view.
Dmap using camera 3 achieves slightly better performance than using multi-cameras (Dmap weighted, CSR-net backnoned) because most people can already be seen in camera 3. This also suggests that ``Dmap weighted'' cannot fuse the multi-view information well. Nonetheless, our fusion methods all outperform the Dmap weighted, demonstrating the efficacy of the projection and fusion stages.
\zqq{The feature concatenation method outputs a scene-level count, and does not consider the geometry relationship of the camera views, thus achieving worse results than the proposed methods. The Stitching method, which only considers the distance between objects and cameras but neglects the occlusions in the camera views, also performs worse than our methods.}
Finally, using multiple cameras to count with ``Detection+ReID'' improves the performance over single cameras, but still has higher error than our fusion methods.

\textbf{PETS S1L1.}
We next compare our method with a traditional multi-view counting method ``Hybrid" \citep{dittrich2017people} on the subset of PETS2009 dataset, PETS S1L1, which is presented in Table \ref{table:comprison_traditional}. \cite{dittrich2017people} proposed two approaches (head detector and count regression) by fusing hand-crafted features (corner points or Harr feature) from multiple cameras for multi-view counting. Similar to \cite{dittrich2017people}, we use PETS2009 S1L1 13\_57 (view 1 and 2) for training and 13\_59 (view 1 and 2) for testing. Our fusion models (FCN-7 backbone) all achieve better performance than the multi-view counting method based on traditional hand-crafted low-level features, and MVMSR achieves the best scene-level counting performance.

%


\subsubsection{DukeMTMC}
\par 
The scene-level counting results on DukeMTMC are shown in  Table \ref{tab:scene_count} (middle row) and Table \ref{table:DukeMTMC_results} (``Scene'' column).
On DukeMTMC, our multi-view fusion models (use CSR-net as backbone) achieve better performance than comparison methods at the scene-level counting task. Due to lower crowd numbers in DukeMTMC, the performance gap among the 3 fusion models is not large -- but MVMS and MVMSR still perform best and MVMSR is better than MVMS. Furthermore, results from comparison methods also show that using a single camera is not adequate for the scene-level counting task. Dmap with a single camera 8 performs slightly better than using multi-cameras (Dmap weighted) due to the large field-of-view of camera 8, and the limitations of the weighted fusion.
\zqq{Since camera views in DukeMTMC dataset share smaller area of overlapping regions and the occlusion issue is not severe, the Stitching method performs relatively better than other comparison methods, but the proposed MVMS and MVMSR still perform better.}



\textbf{DukeMTMC Test Hard.}
Finally, the scene-level counting results on the DukeMTMC Test Hard set, which contains more crowds, are presented in Table \ref{table:DukeMTMC_Test_Hard_set}.
Our fusion model (FCN-7 backbone) achieves better scene-level counting results than the baselines. Among our methods, late fusion has slightly lower error than MVMS/MVMSR.

\subsubsection{CityStreet}

\begin{table}[t]
\centering
\scriptsize
\begin{tabular}{@{}ll|ccc@{}}
    \hline
   Dataset & Method	         & MSE	& NAE & MAE \\
   \hline

   \multirow{4}{*}{PETS2009}
   & CVF \citep{zheng2021learning}	             & -    & - & \textbf{3.08}		\\
   & CVCS	  \citep{zhang2021cross}            & -    & 0.165 & 5.17		\\
   &MVMS             & \textbf{4.83}	& \textbf{0.124} & 3.49		\\
   &MVMSR           & 4.93	& 0.130 & 3.62	\\

   \hline
   \multirow{4}{*}{DukeMTMC}
   & CVF \citep{zheng2021learning}	             & -    & - & \textbf{0.87}	\\
   & CVCS	  \citep{zhang2021cross}            & -    & 0.525 & 2.83		\\
   &MVMS	                  & 1.28   	    & 0.122 & 0.95		\\
   &MVMSR                     & \textbf{1.17}	    & \textbf{0.118} & 0.89		\\

   \hline
   \multirow{4}{*}{CityStreet}

   & CVF \citep{zheng2021learning}	             & -    & - & 7.08		\\
   & CVCS	     \citep{zhang2021cross}         & -  & 0.117  & 9.58 		\\
   & MVMS	                  & 9.02	& 0.096 & 7.36		\\
   & MVMSR                    & \textbf{8.49}	& \textbf{0.086} & \textbf{6.98}		\\
   \hline
\end{tabular}
\caption{The scene-level counting performance of different methods on the 3 datasets. MSE, NAE and MAE are used as evaluation metrics. Note that only MAE is provided in CVF \citep{zheng2021learning}, and MAE and NAE are provided in CVCS \citep{zhang2021cross}.
}
\label{tab:scene_count_new}
\end{table}

\par 
The scene-level counting results on CityStreet are shown in \ref{tab:scene_count} (bottom row) Table \ref{table:CityStreet_results} (``Scene'' column). On CityStreet, our multi-view fusion models achieve better results than the comparison methods. Compared to PETS2009, CityStreet has larger crowds and more occlusions and scale variations. Therefore, the performances of the baseline methods decreases significantly, especially Detection+ReID.
\zqq{Due to large camera angle change and severe occlusions in the CityStreet dataset, Feature concatenation and Stitching cannot perform well on the larger and more complicated dataset.}
Our MVMSR model achieves much better performance on CityStreet than all other models. The reason is the 3 views of the CityStreet dataset have larger view angle change than the other two datasets, which can better demonstrate the effectiveness of the rotation selection in the multi-view fusion process.
Furthermore, similar to the other two datasets, using multi-cameras achieves better scene-level counting performance than using a single camera. 



\begin{table*}
\centering
\begin{tabular}{l|c|ccc}
\hline
                                & \multicolumn{4}{c}{PETS2009 \citep{ferryman2009pets2009}}\\
\hline
     Method (camera)                & Scene          & C1                & C2                & C3              \\
\hline
     Dmap (camera 1)                & 13.74/0.413/12.19  & 4.55/0.213/3.96      & -                 & -                  \\
     Dmap (camera 2)                & 13.48/0.404/12.39   & -                 & 9.43/0.309/8.33      & -                   \\
     Dmap (camera 3)                & {7.95/0.239/6.89}    & -         & -    & {6.37/0.201/5.46}         \\
     Dmap weighted (multiview)      & 7.29/0.182/5.62  & {4.02/0.169/3.61} & {4.25/0.136/3.42} & 5.21/0.149/4.23  \\
\hline
     Detection+ReID (camera 1)      & 17.99/0.545/17.24   & 9.75/0.356/8.57       & -         & -                  \\

     Detection+ReID (camera 2)      & 16.27/0.485/15.54   & -         & 12.33/0.393/11.19     & -                 \\

     Detection+ReID (camera 3)      & 16.27/0.503/16.33  & -         & -         & 15.60/0.472/14.59              \\

     Detection+ReID (multiview)     & {7.01/0.174/5.46}   & {8.22/0.238/6.55}      & {9.33/0.300/7.09}      & {13.50/0.400/11.76}         \\
\hline
     Late fusion (multiview)       & 5.18/0.138/3.92     & 3.20/0.143/2.62      & 4.19/0.137/3.17      & 5.00/0.150/3.97       \\
     Na\"ive (multiview)           &  6.76/0.199/5.42     & 3.13/0.124/2.37       & 5.76/0.179/4.27      & 6.36/0.192/4.92          \\
     MVMS (multiview)          & \textbf{4.83/0.124/3.49} &2.22/0.084/1.66  & 3.67/0.103/2.58 & \textbf{4.58/0.127/3.46 }    \\
     MVMSR (multiview)         & 4.93/0.130/3.62       & \textbf{2.17}/\textbf{0.077}/\textbf{1.57}      & \textbf{3.30}/\textbf{0.097}/\textbf{2.38} & 4.76/0.133/3.64 \\

\hline
\end{tabular}

\caption{
Comparison of the scene-level (left) and the single-view counting (right) measured with mean square error, mean absolute error and relative mean absolute error (MSE/NAE/MAE) on PETS2009. Column ``Scene'' denotes the scene-level counting error. Columns ``C1'', ``C2'' and ``C3'' refer to the single-view counting error for the region within the field-of-view of cameras 1, 2 and 3. ``camera'' indicates the camera(s) used for counting. The late fusion model uses projection normalization, and MVMS and MVMSR uses learnable scale selection and FCN-7 is used as the feature backbone.
}
\vspace{-0.3cm}
\label{table:PETS2009_results}
\end{table*}

\begin{table*}
\centering
\begin{tabular}{l|c|cccc}
\hline
      & \multicolumn{5}{c}{DukeMTMC \citep{ristani2016MTMC}}\\
\hline
    Method (camera)            & Scene         & C2         & C3         & C5         & C8        \\
\hline
    Dmap (camera 2)            & 6.07/0.613/5.19   & 0.97/0.487/0.73   & -          & -          & -                \\
    Dmap (camera 3)            & 8.73/1.000/8.03   & -          & 1.28/0.647/0.79       & -          & -              \\
    Dmap (camera 5)            & 7.39/0.830/6.72   & -          & -          & 0.81/0.575/0.49       & -              \\
    Dmap (camera 8)            & {2.43/0.258/1.87} & -          & -          & -          & 1.57/0.232/1.21         \\
    Dmap weighted  & 2.71/0.250/2.11  & 1.35/0.426/1.02  & {1.28/0.647/0.79}       & {1.42/0.663/0.89}   &{1.83/0.201/1.30}                  \\
\hline
    Det+ReID (camera 2)    & 4.42/0.425/3.51  & 2.28/1.03/2.06       & -         & -         & -                     \\
    Det+ReID (camera 3)    & 7.93/0.890/7.20  & -         & 0.55/\textbf{0.132}/0.25      & -         & -       \\
    Det+ReID (camera 5)   & 7.11/0.782/6.38 & -         & -         & {1.29/0.524/0.96}      & -          \\
    Det+ReID (camera 8)   & 5.85/0.620/5.10  & -         & -         & -         & 4.28/0.541/3.58        \\
    Det+ReID (multiview)  & {2.30/0.355/1.89} & {0.94/0.513/0.75}      & 0.71/0.584/0.40      & 2.37/1.09/1.89    & 3.62/0.422/2.86          \\
\hline
    Late fusion (multiview)    & 1.63/0.187/1.29   & 0.64/0.263/0.45  & 0.56/1.040/0.35   & 0.66/0.548/0.37      & 1.48/0.198/1.15      \\
    Na\"ive (multiview)    & 1.90/0.199/1.47     & 0.59/0.265/0.44      & 0.66/0.970/0.44       & 0.91/0.768/0.60     & 1.64/0.195/1.23               \\
   MVMS (multiview)     & 1.28/0.122/0.95		& 0.50/0.199/0.33 & 0.40/0.677/0.21 & 0.59/\textbf{0.401}/0.31 & 1.11/0.125/0.81\\
   MVMSR (multiview)     & \textbf{1.17}/\textbf{0.118}/\textbf{0.89}  & \textbf{0.45}/\textbf{0.183}/\textbf{0.30} & \textbf{0.38}/0.758/\textbf{0.19}  & \textbf{0.54}/0.410/\textbf{0.29}   & \textbf{1.02}/\textbf{0.118}/\textbf{0.76}       \\
\hline
\end{tabular}

\caption{
Comparison of the scene-level (left) and the single-view counting (right) using mean square error, mean absolute error and relative mean absolute error (MSE/NAE/MAE) on DukeMTMC.  CSR-net is used as the feature backbone. See the caption of Table \ref{table:PETS2009_results} for further description.
}
\vspace{-0.3cm}

\label{table:DukeMTMC_results}
\end{table*}

\subsubsection{\zqqblue{Comparison with concurrent methods}}
\zqqblue{We next compare with two recent multi-view counting methods published concurrently
during the revision of our paper:
CVF \citep{zheng2021learning}  and
CVCS \citep{zhang2021cross}.
The comparison is shown in Table \ref{tab:scene_count_new}, and our proposed method
achieves the best performance on the largest dataset CityStreet.}

\subsection{Single-view counting performance}

\par 
We next evaluate the single-view counting performance, which is the people count within the single-camera's field-of-view. Here we mainly aim to show that multi-view information can improve single-view counting over using a single camera. Note, the comparison method feature concatenate and stitching directly output a scene-level count and are not used for comparison in this section.


\sssection{PETS2009:} The single-view counting results on PETS2009 are shown in Table \ref{table:PETS2009_results}. Columns `C1', `C2', and `C3' correspond for single-view counting in regions 
within the fields-of-view of cameras 1, 2, and 3, respectively. On PETS2009, our 3 multi-view fusion models can achieve better results than the two comparison methods in terms of all single-camera counting, which demonstrates that the proposed multi-view fusion DNNs can well integrate the information from multi-views to improve the counting performance in different regions. Furthermore, MVMS performs much better than other models, which also shows the multi-scale framework with scale selection strategies can improve the feature-level fusion to achieve better performance. Finally, the comparison methods' single-view counting performance can also be improved with the aid of other cameras.

\begin{table*}
\centering
\begin{tabular}{l|c|ccc}
\hline
    Dataset &  \multicolumn{4}{c}{CityStreet}\\
\hline
    Method (camera)         & Scene    & C1        & C3        & C4        \\
\hline
    Dmap (camera 1)         & 11.70/0.110/9.31    & 10.21/0.112/8.51     & -         & -   \\
    Dmap (camera 3)         & 11.64/0.199/9.41    & -         & 11.83/0.129/9.23     & -         \\
    Dmap (camera 4)          & 24.24/0.256/21.92   & -         & -         & 22.84/0.240/20.30     \\
    Dmap weighted (multiview) & 11.46/0.120/9.36  &{9.30/0.101/7.87}       & {11.19/0.121/9.19}     & {12.84/0.116/10.16}     \\
\hline
    Detection+ReID (camera 1)   & 49.68/0.513/45.80   & 45.71/0.483/41.38     & -         & -         \\
    Detection+ReID (camera 3)   &  45.09/0.453/40.87  & -         & 37.94/0.391/32.94     & -         \\
    Detection+ReID (camera 4)   & 35.10/0.323/30.03  & -         & -         & 33.16/0.311/28.57     \\
    Detection+ReID (multiview)  & {21.18/0.193/17.48}  & {42.36/0.456/37.76} & {35.10/0.355/29.27}    & {22.84/0.228/18.21} \\

\hline
    Late fusion              & 9.63/0.099/8.06	& 9.71/0.110/8.36 &9.51/0.110/7.99 &9.01/0.089/7.46 	\\
    Na\"ive early fusion     & 9.85/0.100/8.11	& 10.04/0.108/8.35 &9.42/0.106/7.74 &9.65/0.098/7.94	\\


    MVMS	(multiview)      & 9.02/0.096/7.36	& 	9.59/0.110/7.87  &8.44/0.100/6.87 & \textbf{7.59}/0.081/\textbf{6.24}   \\
    MVMSR (multiview)        & \textbf{8.49}/\textbf{0.086}/\textbf{6.98}	& \textbf{8.51}/\textbf{0.094}/\textbf{7.05} &\textbf{8.06}/\textbf{0.089}/\textbf{6.49} &7.89/\textbf{0.078}/6.44 	\\
\hline
\end{tabular}

\caption{
Comparison of the scene-level and the single-view counting using mean square error, mean absolute error and relative mean absolute error (MSE/NAE/MAE) on CityStreet.  CSR-net is used as the feature backbone. See the caption of Table \ref{table:PETS2009_results} for further description.
}
\label{table:CityStreet_results}
\end{table*}

\begin{table*}[t]
\centering
\begin{tabular}{ll|c|c|c|c|c}
    \hline
   Backbone & Method	         & MSE	& NAE & MAE/GAME(0) & GAME(1) & GAME(2) \\
   \hline
   \multirow{4}{*}{FCN-7}
   &Late fusion              & 10.24	& 0.097 & 8.12	& \textbf{13.05} & 21.14	\\
   &Na\"ive early fusion     & 10.11	& 0.096 & 8.10	& 13.06 & \textbf{20.98}	\\
   &MVMS	                 & 10.05	& 0.096 & 8.01	& 13.67 & 21.99	\\
   &MVMSR                    & \textbf{9.73}	& \textbf{0.090} & \textbf{7.37}	& 13.89 & 22.60	\\
   \hline
   \multirow{ 4}{*}{CSR-net}
   &Late fusion              & 9.63	& 0.099 & 8.06	& 12.75 & 23.10	\\
   &Na\"ive early fusion     & 9.85	& 0.100 & 8.11	& 12.73 & 22.93	\\
   &MVMS	                 & 9.02	& 0.096 &7.36 	& 11.95 & 20.44	\\
   &MVMSR                    & \textbf{8.49}	& \textbf{0.086} & \textbf{6.98}	& \textbf{11.39} & \textbf{19.79}	\\
   \hline
   \multirow{4}{*}{LCC}
   &Late fusion              & 10.51	&0.113  & 8.71	& 15.45 & 26.29	\\
   &Na\"ive early fusion     & 9.96	   & 0.103 & 7.97	& 12.99 & 22.56	\\
   &MVMS	                 & 9.86   	& 0.093 & 7.67	& 13.92 & 22.97	\\
   &MVMSR                    & \textbf{9.46}	& \textbf{0.086} & \textbf{7.42}	& \textbf{12.77} & \textbf{21.22}	\\
   \hline
\end{tabular}
\caption{Comparison of different backbones for scene-level counting performance on CityStreet, where the settings for the rotation module are the same: the filter number is 32, the layer number is 3 and the quantization angle is $45^{\circ}$.}
\label{tab:backbone}

\vspace{-0.5cm}

\end{table*}

\sssection{DukeMTMC:} 
On DukeMTMC, our multi-view fusion models can achieve better performance than comparison methods on most camera-views (see Table \ref{table:DukeMTMC_results}). Detection+ReID achieves the good result on camera 3 because this camera is almost parallel to the horizontal plane, has low people count, and rarely has occlusions, which is an ideal operating regime for the detector. Finally, the single-view counting performance is mostly improved with the aid of multi-cameras.

\sssection{CityStreet:}
On CityStreet (see Table \ref{table:CityStreet_results}), our 3 multi-view fusion models achieve better results than the comparison methods. Due to severe occlusions and scale changes, the Detection+ReID methods perform badly, even with the aid of other cameras. However, using multi-cameras still improves the single-view counting performance, and our methods are the most effective at multi-view fusion.

\begin{table*}[t]
\centering
\begin{tabular}{ll|c|c|c|c|c}
    \hline
   Backbone & Method	         & MSE	& NAE & MAE/GAME(0) & GAME(1) & GAME(2) \\
   \hline
   \multirow{6}{*}{PETS2009}

   &MVMS(FCN-7)   & \textbf{4.83}	& \textbf{0.124} & \textbf{3.49}	& \textbf{5.30} &\textbf{ 6.27}	\\
   &MVMS(CSR-net)  & 5.26 & 0.140 &3.99 &6.27 &7.51   	\\
   &MVMS (LCC)	    &5.62 &0.151 &4.36 & 6.68& 7.98    	\\
   &MVMSR (FCN-7)  & 4.93	& 0.130 & 3.62	& 5.37 & 6.98	\\
   &MVMSR (CSR-net) & 5.51 &0.140 &4.15 &6.56 & 8.30      	\\
   &MVMSR (LCC)	    &5.60 &0.151 & 4.37 & 5.80 & 6.94      	\\

   \hline
   \multirow{6}{*}{DukeMTMC}
   &MVMS(FCN-7)      & 1.24   	    & 0.170 & 1.03	& 1.53 & 1.92	\\
   &MVMS(CSR-net)     & 1.28	& 0.122 & 0.95	& 1.24 & 1.50	\\
   &MVMS (LCC)	    & 1.38	& 0.132 & 1.04	& 1.26 & 1.49	\\
   &MVMSR (FCN-7)   & 1.31	    & 0.144 & 1.01	& 1.50 & 2.02	\\
   &MVMSR (CSR-net) & \textbf{1.17}	& \textbf{0.118} & \textbf{0.89}	&\textbf{1.19 }&\textbf{1.42}	\\
   &MVMSR (LCC)	    & 1.26	& 0.129 & 0.94	& 1.29 & 1.57	\\

   \hline
   \multirow{6}{*}{CityStreet}
   &MVMS(FCN-7)              & 10.05	& 0.096 & 8.01	& 13.67 & 21.99	\\
   &MVMS(CSR-net)     & 9.02	& 0.096 & 7.36	& 11.95 & 20.44	\\
   &MVMS (LCC)	                 & 9.86	& 0.093 & 7.67	& 13.92 & 22.97	\\
   &MVMSR (FCN-7)                & 9.73	& 0.090 & 7.37	& 13.89 & 22.60	\\
   &MVMSR (CSR-net)       & \textbf{8.49}	& \textbf{0.086} & \textbf{6.98}	& \textbf{11.39} & \textbf{19.79}	\\
   &MVMSR (LCC)	                   & 9.46	& 0.086 & 7.42	& 12.77 & 21.22	\\

   \hline
\end{tabular}
\caption{Comparison of different backbones for MVMS/MVMSR on PETS2009, DukeMTMC and CityStreet. For MVMSR, the filter number is 32, the layer number is 3 and the quantization angle is $10^{\circ}$, $45^{\circ}$ and $45^{\circ}$ for PETS2009, DukeMTMC and CityStreet, respectively.}
\label{tab:backbone2}
\end{table*}

\subsection{Ablation studies}

We next present ablation studies on the various components of our fusion pipeline.

\begin{table*}
\centering
\begin{tabular}{l|c@{\hspace{0.25cm}}c@{\hspace{0.25cm}}c@{\hspace{0.25cm}}c|c@{\hspace{0.25cm}}c@{\hspace{0.25cm}}c@{\hspace{0.25cm}}c@{\hspace{0.25cm}}c|c@{\hspace{0.25cm}}c@{\hspace{0.25cm}}c@{\hspace{0.25cm}}c}
\hline  
    Dataset & \multicolumn{4}{c|}{PETS2009}       & \multicolumn{5}{c|}{DukeMTMC}           & \multicolumn{4}{c}{CityStreet}\\
\hline
    Region  & C1 & C2 & C3 & Scene                           & C2 & C3 & C5 & C8 & Scene         & C1 & C3 & C4 & Scene  \\
\hline
   Late fusion (w/ PN)     & \textbf{2.62} & \textbf{3.17}  & \textbf{3.97} & \textbf{3.92}
             & \textbf{0.49} & 0.77  & \textbf{0.39}   & \textbf{1.15}  & \textbf{1.27}
             & \textbf{8.14} & \textbf{7.72}  & \textbf{8.08} & \textbf{8.12}    \\

  Late fusion (w/o PN) & 2.75 & 3.86  & 4.37   & 4.22
            & 0.63 & \textbf{0.73}  & 0.51   & 1.31  & 1.43
            &  9.89 & 9.60  & 9.82   & 9.87   \\
\hline
    MVMS (fixed)    & 1.74 & \textbf{2.57}  & 3.81   & 3.82
             & 0.65 & \textbf{\textbf{0.46}}  & \textbf{0.88}   & 1.44  & 1.09
             & 8.11  & 7.83   &8.32 & \textbf{7.80}     \\

   MVMS (learnable)  & \textbf{1.66} & 2.58  & \textbf{3.46}   & \textbf{3.49}
               & \textbf{0.63} & 0.52  & 0.94   & \textbf{1.36}  & \textbf{1.03}
                & \textbf{7.99} & \textbf{7.63}  & \textbf{7.91}   & 8.01  \\
   \hline
\end{tabular}
\caption{Ablation study (MAE) comparing the late fusion model with and without projection normalization (PN), and MVMS with fixed or learnable scale selection.}
\label{table:Normalization_study}
\vspace{-0.5cm}
\end{table*}
\subsubsection{Backbone for MVMS and MVMSR}
\label{text:CSRnet}


We compare different feature extraction backbones in the proposed multi-view counting framework: FCN-7, CSR-net \citep{li2018csrnet} (first 7 layers of VGG) and \zqq{LCC \citep{liu2020adaptive} (use the output of the scale-aware module as extracted features)}.

First, the counting results of the 4 fusion models on CityStreet are presented in Table \ref{tab:backbone}. Generally, using larger backbone performs better than FCN-7.
However, the performance gap using different is larger for Dmap weighted, indicating that the single-view density maps of CSR-Net are more accurate.
\nabc{This suggests that the scale-selection module in MVMS is sufficient for handling scale changes in multi-view counting, and the benefits of using dilated convolutions to handle single-view scale changes are diminished.}
\zqq{Furthermore, the improvement of MVMSR and MVMS over late/na\"ive fusion is consistent among the 3  backbones. Finally, the proposed MVMSR method achieves the best performance on all 3 backbones, which indicates the effectiveness of the rotation selection module.}

\zqq{
Second, the counting results of MVMS and MVMSR on the 3 datasets are presented in Table \ref{tab:backbone2}.
Generally, with larger backbone, the performance of MVMS or MVMSR can be improved on the CityStreet and DukeMTMC dataset, and MVMSR is better than MVMS. On PETS2009 dataset, the smaller backbone performs better than larger backbones, and the possible reason is the larger backbone models are overfitting on the dataset since more CNNs layers are used. Furthermore, on PETS2009, since the camera angle change is not large enough and most people can be seen by camera 3, the best performance of MVMSR is slightly worse than MVMS.  Nonetheless, MVMSR is better than MVMS on the more complicated datasets, CityStreet and DukeMTMC.
}

\subsubsection{Normalization in the late fusion model}
\par We perform an ablation study on the late fusion model (FCN-7 backbone) with and without the projection normalization step, and the results are presented in Table \ref{table:Normalization_study} (top). Using projection normalization reduces the error of the late fusion model, compared to not using the normalization step.
\nnabc{This demonstrates the importance of maintaining the total count when projecting the density map into the ground-plane representation.}

\subsubsection{Scale selection in MVMS}
\par We perform an ablation study on the scale-selection strategy of MVMS, and the results are presented in Table \ref{table:Normalization_study} (bottom). Most of the time the learnable scale-selection strategy can achieve lower error than fixed scale-selection. We note that using MVMS with fixed scale-selection strategy still outperforms the na\"ive early fusion, which performs no scale selection. Thus obtaining features that have consistent scales across views is an important step when fusing the multi-view feature maps.

\par
The proposed two scale selection modules use different mask methods: discrete mask for fixed scale selection and soft-mask for learnable scale selection.
We perform another ablation study on fixed scale selection using the soft-mask, which is presented in Table \ref{tab:fixed_soft}. When using soft-masks, learnable scale selection still outperforms fixed scale selection.

\begin{table}[t]
\centering
\begin{tabular}{cccc}
    \hline
  Dataset		& fixed discrete	& fixed soft  & learnable soft \\
   \hline
   PETS2009			& 3.82	& 3.59	& 3.49	 \\
   CityStreet		& 7.80	& 8.55	& 8.01	 \\
   \hline
\end{tabular}
\caption{MAE comparison of MVMS model (FCN-7 backbone) selection module settings: fixed scale selection with discrete mask or soft-mask, and learnable scale selection with soft-mask.
}
\vspace{-0.5cm}
\label{tab:fixed_soft}
\end{table}

\subsubsection{\add{Rotation module in MVMSR}}

\zqq{We perform the ablation study on the rotation module of MVMSR on the CityStreet dataset (with CSR-net backbone), including the number of filters $F$, the number of layers  $L$, and the rotation quantization angle $Q$ of the rotation selection layer.}

\zqq{
\textbf{Number of filters $F$:} The results of the ablation study on the number of filters $F$ is shown in Table \ref{tab:filter_num}, where the number of layers in rotation selection  is 3 and the rotation quantization angle is $45^{\circ}$.
Increasing the number of  filters 
does not necessarily improve the performance of the scene-level counting performance. The reason is because the rotation selection layer
naturally handles the effect of feature rotations in the projection step, and thus less filters are required when compared to without the rotation selection, where each rotation needs a separate filter. Besides, using more filters decreases the speed of the model in the inference stage. Therefore, we choose $F=32$ in the remaining experiments, whose performance is better compared to others.
}

\zqq{
\textbf{Number of layers $L$:} We perform an ablation study on the number of layers $L$ of the rotation module in Table \ref{tab:layer_num}, where the number of filters is 32 and the rotation quantization angle is $45^{\circ}$.
From the table, we conclude that fewer rotation selection layers may not reduce the feature rotation effect and too many layers may make the model overfit and decrease the counting performance.
The choice of $L=5$ achieves the best performance in terms of MAE, MSE and NAE metric while $L=3$ achieves the best performance in terms of GAME(1) and GAME(2).  We use $L=5$ in the remaining experiments.
}

\zqq{
\textbf{Quantization angle $Q$:}}
Finally, we perform an ablation study on the quantization angle $Q$ in the rotation selection module of MVMSR on CityStreet, and the results are presented in Table \ref{tab:quantization_angle_q}. The choice of the quantization angle $Q$ involves the balance between the benefit of rotation selection and the extra learning complexity caused by the multi-rotations of the features. $Q = 45^{\circ}$ has the best performance, compared to larger and smaller quantization angles on CityStreet.

\zqq{In the main experiments tables (Table \ref{tab:scene_count} and \ref{table:CityStreet_results}), we report the result on CityStreet with $F=32$, $L=3$ and $Q = 45^{\circ}$ for better overall performance in terms of all evaluation metrics than the comparison methods.}

\begin{table}[t]
\centering
\begin{tabular}{ccccccc}
    \hline
    $F $		&MAE & MSE	& NAE &G(1) &G(2) & FPS   \\
   \hline
    8		& 7.56	& 9.36	& 0.091	 & 11.54	& 20.70 & \textbf{5.8}		\\
    16		& 7.24	& 8.74	& 0.088	 & 11.73	& 19.81 & 5.6		\\
    32		& \textbf{6.98}	& \textbf{8.49}	& \textbf{0.086}	& \textbf{11.39}	& \textbf{19.79}  & 5.5		\\
    64		& 7.09	& 8.73	& 0.088	 & 11.90	& 20.47 & 3.4		\\
    128		& 7.08	& 8.82	& 0.092	 & 12.31	& 21.28 & 0.6	\\
   \hline
\end{tabular}
\caption{The ablation study on filter number $F$  of the rotation module in MVMSR on CityStreet. Here $L=3$ and $Q=45$. G(1) and G(2) are GAME(1) and GAME(2), respectively.}
\label{tab:filter_num}
\end{table}

\begin{table}[t]
\centering
\begin{tabular}{cccccc c}
    \hline
    $L $		&MAE & MSE	& NAE &G(1) &G(2) & FPS   \\
   \hline
    1		& 7.32	& 8.81	& 0.090	&12.27 & 20.90 & \textbf{6.8}		\\
    3		& 6.98	& 8.49	& 0.086	 &\textbf{11.39} &\textbf{19.79} & 5.5		\\
    5		& \textbf{6.63}	& \textbf{8.25}	& \textbf{0.082} & 12.11 & 20.70	 & 4.5		\\
    7		& 6.77	& 8.46	& 0.086	 & 11.45 &20.40 & 3.8		\\
    9		& 7.00	& 8.77	& 0.082	& 11.56 &20.24 & 3.3	\\
   \hline
\end{tabular}
\caption{The ablation study on layer number $L$ of the rotation module in MVMSR on CityStreet. Here $F=32$ and $Q=45^{\circ}$.}
\label{tab:layer_num}
\end{table}

\begin{table}[t]
\centering
\begin{tabular}{ccccccc}
    \hline
    $Q $		&MAE & MSE	& NAE & G(1) &G(2) & FPS   \\
   \hline
    $15^{\circ}$		& 7.48	& 9.04	& 0.092	&12.68 &21.60 & 2.0		\\
    $30^{\circ}$		& 7.26	& 9.02	& 0.085	&\textbf{11.06} &\textbf{19.34} & 3.5		\\
    $45^{\circ}$		& \textbf{6.63}	& \textbf{8.25}	& \textbf{0.082} &12.11 &20.70	 & 4.5		\\
    $60^{\circ}$		& 7.02	& 8.56	& 0.084	&11.54 &20.19  & 5.2		\\
    $75^{\circ}$	    & 7.01	& 8.47	& 0.089	&11.96 &20.65  & 5.8	\\
    $90^{\circ}$	    & 7.20	& 8.63	& 0.088	&11.96 & 20.17 & \textbf{6.1}	\\

   \hline
\end{tabular}
\caption{The ablation study on rotation quantization angle $Q$ of the rotation module in MVMSR on CityStreet. Here $F=32$ and $L=5$.}
\label{tab:quantization_angle_q}

\end{table}

\subsubsection{Detection+ReID with Detection or ReID ground-truth}
\par 
In crowd scenes, detection methods are limited by severe occlusions among the crowd, while ReID methods are hindered by detection errors, partial occlusions, scale changes between cameras, and low image-patch resolution.
To illustrate the difficulties, we use the ground-truth inter-camera associations (\emph{i.e.}, the best possible ReID) on the people detections and get counting MAE 30.3 on CityStreet, which is worse than our density-map fusion methods.
Likewise, we apply ReID (Circle2021) on the ground-truth person boxes (\emph{i.e.}, the best possible detector), and get counting MAE 10.7. Integrating multi-view detection and ReID for multi-view crowd counting would be interesting future work, and our dataset could serve as a test-bed.

\begin{table}
\centering
\small
\begin{tabular}{l|c}
\hline
    Method                                     & FPS \\
\hline
   Dmap weighted (FCN-7)                              & 58.5     \\
   Dmap weighted (CSR-net)                            & 9.4     \\
   Detection+ReID                              & 0.3   \\
   Feature concatenation                       & 3.4   \\
   Stitching                                   & 8.7   \\
\hline
   Late fusion (FCN-7)                         & 27.8      \\
   Na\"ive early fusion (FCN-7)                & 30.7    \\
   MVMS (FCN-7)                                & 19.8    \\
   MVMSR (FCN-7)                               & 11.1    \\
\hline
   Late fusion (CSR-net)                        & 5.2      \\
   Na\"ive early fusion (CSR-net)               & 8.5    \\
   MVMS (CSR-net)                               & 7.9   \\
   MVMSR (CSR-net)                              & 5.5   \\
\hline
   Late fusion (LCC)                           & 3.9      \\
   Na\"ive early fusion (LCC)                  & 5.0    \\
   MVMS (LCC)                                  & 4.4    \\
   MVMSR (LCC)                                 & 0.5    \\
\hline

\end{tabular}
\caption{Running speed comparison on the CityStreet dataset.}
\label{table:speed}
\end{table}

\subsubsection{Running speed comparison}

\zqq{We compare the running speed of different methods on the CityStreet dataset in the Table \ref{table:speed}.
Times are recorded for an Intel Xeon CPU E5-2543@3.30GHz with a Nvidia Geforce GTX 1080 Ti GPU.
Generally, larger or deeper backbones (CSR-net and LCC) have lower running speed than lighter backbones (FCN-7).
Comparing our methods, na\"{i}ve early fusion is faster than late fusion because only one decoder module is needed to predict the scene-level density map, whereas late fusion has additional computations for predicting the density maps for each camera-view. MVMS is slower than na\"{i}ve early fusion because MVMS extracts multi-scale features, compared to a single feature scale for na\"{i}ve early fusion. Finally,  MVMSR is slower than MVMS due to the additional network depth from the rotation module.
The comparison method Dmap weighted (FCN-7) is faster than Dmap weighted (CSR-net) and Dmap\_weighted (CSR-net), due its smaller backbone.
For Detection+ReID (Circle2020), it uses large ReID models, so the running speeds is slow.
The Stitching method's speed is similar to Dmap weighted (CSR-net), due to the same backbone model.
For feature concatenation method, a large CNN model is used, so the running speed is comparable to our method's speed with larger backbones.
}

\section{Discussion and Conclusion}
\par In this paper, we propose a DNNs-based multi-view counting framework that fuses camera-views to predict scene-level ground-plane density maps for wide-area crowd counting. Both late fusion of density maps and early fusion of feature maps are studied. For late fusion, a projection normalization method is proposed to counter the effects of stretching caused by the projection operation. For early fusion, a multi-scale approach is proposed that selects features that have consistent scales across views. We also propose a rotation selection module to handle rotated features introduced by the projection operation. To advance research in multi-view counting, we collect a new dataset of large scene containing a street intersection with large crowds.
\zqq{ From the experiment results, our methods' performance
gain over other comparison methods are larger on CityStreet, which is a larger and crowded scene. On the other hand, when the scene is not crowded enough, such as
DukeMTMC, other methods can also achieve good performance. Nonetheless, our methods MVMS/MVMSR achieve the best performance on all 3 datasets.}


\zqq{In this paper, we focus on multi-camera counting when camera calibrations are known (like many other multi-camera vision tasks, such as 3D human pose estimation and multi-camera detection and tracking). The situation when the surveillance cameras orientation or intrinsic parameters change gradually is interesting future work. One way to handle this situation would be to build an automatic calibration system, such as AutoClib \citep{bhardwaj2018autocalib}, which uses car type and size to help calibrate the cameras, or \citep{ammar2019geometric}, which computes a homography matrix  for transforming the image to a geometrically correct bird's eye (overhead) view. Other calibration methods from 3D reconstruction \citep{agarwal2011building,snavely2006photo} could also be used.}

Besides, adapting our framework to moving cameras and unknown camera parameters (using the full spatial transformer net) is interesting future work.
In addition, we have trained and tested  the network on each dataset individually. Another interesting future direction is on {\em cross-scene} multi-view counting, where the scenes in the test set are distinct from those in the training set -- however, this requires more multi-view scenes to be collected.
\zqq{Since collecting videos of real scenes is difficult, especially during the pandemic, one recent work \citep{zhang2021cross} has collected a synthetic dataset for cross-scene cross-view multi-view counting.}

\begin{acknowledgements}
This work was supported by grants from the Research Grants Council of the Hong Kong Special Administrative
Region, China (Project No. [T32-101/15-R] and CityU 11212518), and by a Strategic Research Grant from City
University of Hong Kong (Project No. 7004887). We are grateful for the support of NVIDIA Corporation with the
donation of the Tesla GPU used for this research.
\end{acknowledgements}

\bibliographystyle{spbasic}      
\bibliography{egbib}   

\end{document}